\documentclass[dvipsnames]{article} %
\usepackage{colm2024_conference}

\usepackage{booktabs}
\usepackage{enumitem}
\usepackage{wrapfig}
\usepackage{algorithm}
\usepackage{algpseudocode}
\usepackage{graphicx}
\usepackage[misc]{ifsym}

\usepackage{microtype}
\usepackage{amsmath}
\usepackage{colortbl}
\usepackage[utf8]{inputenc}
\usepackage[T1]{fontenc}
\definecolor{lightgray}{rgb}{0.9,0.9,0.9}
\usepackage{caption}
\usepackage{subcaption}
\usepackage{setspace}
\usepackage{url}
\usepackage{multirow}
\usepackage{tabularx}
\usepackage{blindtext}
\usepackage{pgfplots}
\pgfplotsset{compat=1.18} 
\usepackage{tikz}
\usetikzlibrary{er,positioning,bayesnet}
\usepackage{makecell}
\usepackage{tipa}
\usepackage{siunitx}
\usepackage{nicefrac}
\usepackage{listings}
\usepackage[raster,skins, most]{tcolorbox} %
\usepackage{xltabular}
\usepackage{adjustbox}
\usepackage{xurl}
\usepackage{rotating}
\usepackage[normalem]{ulem}

\usepackage{fontawesome}

\usepackage{pifont}    
\usepackage{xcolor}      
\newcommand\inb[1]{\colorbox{gray!20}{\lstinline|#1|}}
\newcommand{\cmark}{\ding{51}}
\newcommand{\xmark}{\ding{55}}

\useunder{\uline}{\ul}{}

\usepackage{amsmath,amsfonts,bm}

\def\eqref#1{equation~\ref{#1}}

\def\1{\bm{1}}

\DeclareMathAlphabet{\mathsfit}{\encodingdefault}{\sfdefault}{m}{sl}
\SetMathAlphabet{\mathsfit}{bold}{\encodingdefault}{\sfdefault}{bx}{n}

\renewcommand{\ghlink}{https://github.com/Alibaba-NLP/DeepResearch}

\newcommand*\justify{%
  \fontdimen2\font=0.4em
  \fontdimen3\font=0.2em
  \fontdimen4\font=0.1em
  \fontdimen7\font=0.1em
  \hyphenchar\font=`\-
}

\renewcommand{\texttt}[1]{%
  \begingroup
  \ttfamily
  \begingroup\lccode`~=`/\lowercase{\endgroup\def~}{/\discretionary{}{}{}}%
  \begingroup\lccode`~=`[\lowercase{\endgroup\def~}{[\discretionary{}{}{}}%
  \begingroup\lccode`~=`.\lowercase{\endgroup\def~}{.\discretionary{}{}{}}%
  \catcode`/=\active\catcode`[=\active\catcode`.=\active
  \justify\scantokens{#1\noexpand}%
  \endgroup
}

\usepackage{makecell}
\usetikzlibrary{tikzmark}
\makeatletter
\newcommand*\myfontsize{%
  \@setfontsize\myfontsize{7}{8}%
}
\makeatother

\definecolor{uclablue}{RGB}{159, 195, 224}

\definecolor{uclagold}{RGB}{255, 240, 180}

\definecolor{aliceblue}{RGB}{255, 238, 241}

\definecolor{cadmiumgreen}{rgb}{0.0, 0.42, 0.24}

\definecolor{myred}{rgb}{0.7, 0.3, 0.0}
\definecolor{myblue}{rgb}{0.2, 0.3, 0.6}
\definecolor{babygreen}{rgb}{0.85, 0.97, 0.85}

\definecolor{purple1}{RGB}{126, 107, 196}
\definecolor{purple2}{RGB}{199, 158, 207}
\definecolor{purple3}{RGB}{214, 200, 255}
\definecolor{purple4}{RGB}{254, 240, 255}

\definecolor{deepblue}{RGB}{48, 58, 82}

\newcommand{\symboletongyi}{\raisebox{0pt}{~\includegraphics[scale=0.012]{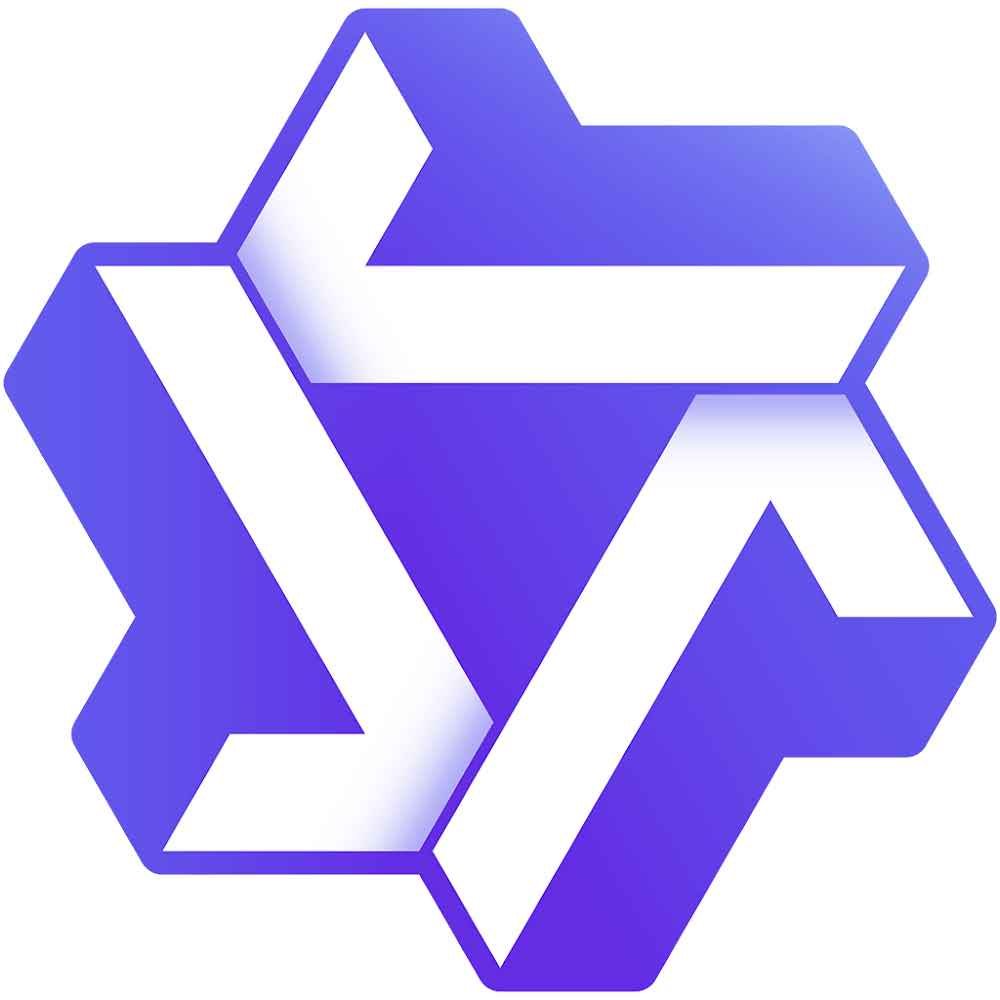}}~}

\definecolor{deepPurple}{HTML}{330066}

\definecolor{uclablue_old}{rgb}{0.15, 0.45, 0.68}
\hypersetup{
    breaklinks,
    citecolor=uclablue_old,
    colorlinks=true,
}

\newtcolorbox{mybox}[2][]
  {colback = black!5!white, colframe = black!75!black, fonttitle = \bfseries,
    colbacktitle = black!100!black, enhanced, before upper={\fontsize{8}{11}\obeyspaces\obeylines\selectfont}, fontupper=\selectfont,
    attach boxed title to top left={yshift=-2.2mm,xshift=4mm},
    title=#2,#1}

\newcommand{\myname}{AgentFrontier}

\title{%
\raisebox{-2.0em}{
  \parbox[t]{0.35in}{\includegraphics[width=0.6in]{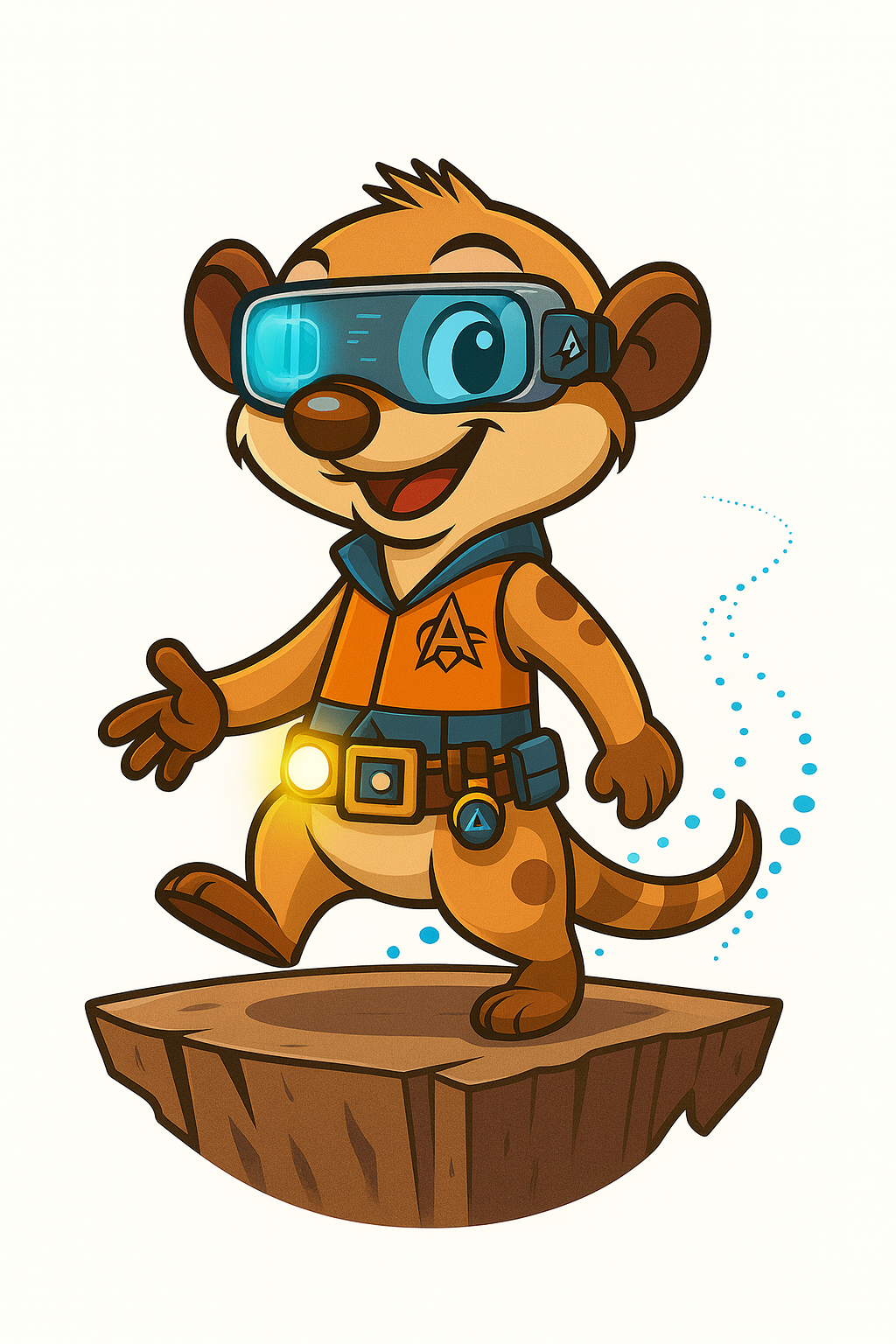}} 
  }
\begin{tabular}[t]{l} 
  \parbox[t]{0.8\textwidth}{\centering 
    \myname: Expanding the Capability Frontier of LLM Agents with ZPD-Guided Data Synthesis
  }
\end{tabular}
}

\author{%
\small{Xuanzhong Chen$^{*}$, Zile Qiao\thanks{Equal Core Contributors. xuanzhchen@gmail.com, qiaozile.qzl@alibaba-inc.com}\hspace{0.5mm} $^{(\textrm{\Letter})}$, Guoxin Chen, Liangcai Su, Zhen Zhang, Xinyu Wang, Pengjun Xie, Fei Huang, Jingren Zhou, Yong Jiang$^{(\textrm{\Letter})}$}%
  \\[1em]               
  {\fontsize{10pt}{11pt}\selectfont          
Tongyi Lab\symboletongyi, Alibaba Group}\\
}

\begin{document}

\maketitle

\begingroup
  \renewcommand\thefootnote{\Letter}  
  \footnotetext{Corresponding author. \{qiaozile.qzl, yongjiang.jy\}@alibaba-inc.com} 
\endgroup

\begin{abstract}
Training large language model agents on tasks at the frontier of their capabilities is key to unlocking advanced reasoning. We introduce a data synthesis approach inspired by the educational theory of the \textit{Zone of Proximal Development} (ZPD), which defines this frontier as tasks an LLM cannot solve alone but can master with guidance. To operationalize this, we present the \textbf{AgentFrontier Engine}, an automated pipeline that synthesizes high-quality, multidisciplinary data situated precisely within the LLM's ZPD. This engine supports both continued pre-training with knowledge-intensive data and targeted post-training on complex reasoning tasks. From the same framework, we derive the \textbf{ZPD Exam}, a dynamic and automated benchmark designed to evaluate agent capabilities on these frontier tasks. We train \textbf{AgentFrontier-30B-A3B} model on our synthesized data, which achieves state-of-the-art results on demanding benchmarks like Humanity's Last Exam, even surpassing some leading proprietary agents. Our work demonstrates that a ZPD-guided approach to data synthesis offers a scalable and effective path toward building more capable LLM agents.
\end{abstract}

\begin{figure*}[h]
\vspace{-10pt}
\centering
\begin{subfigure}[b]{0.48\textwidth}
    \centering
    \includegraphics[width=\linewidth]{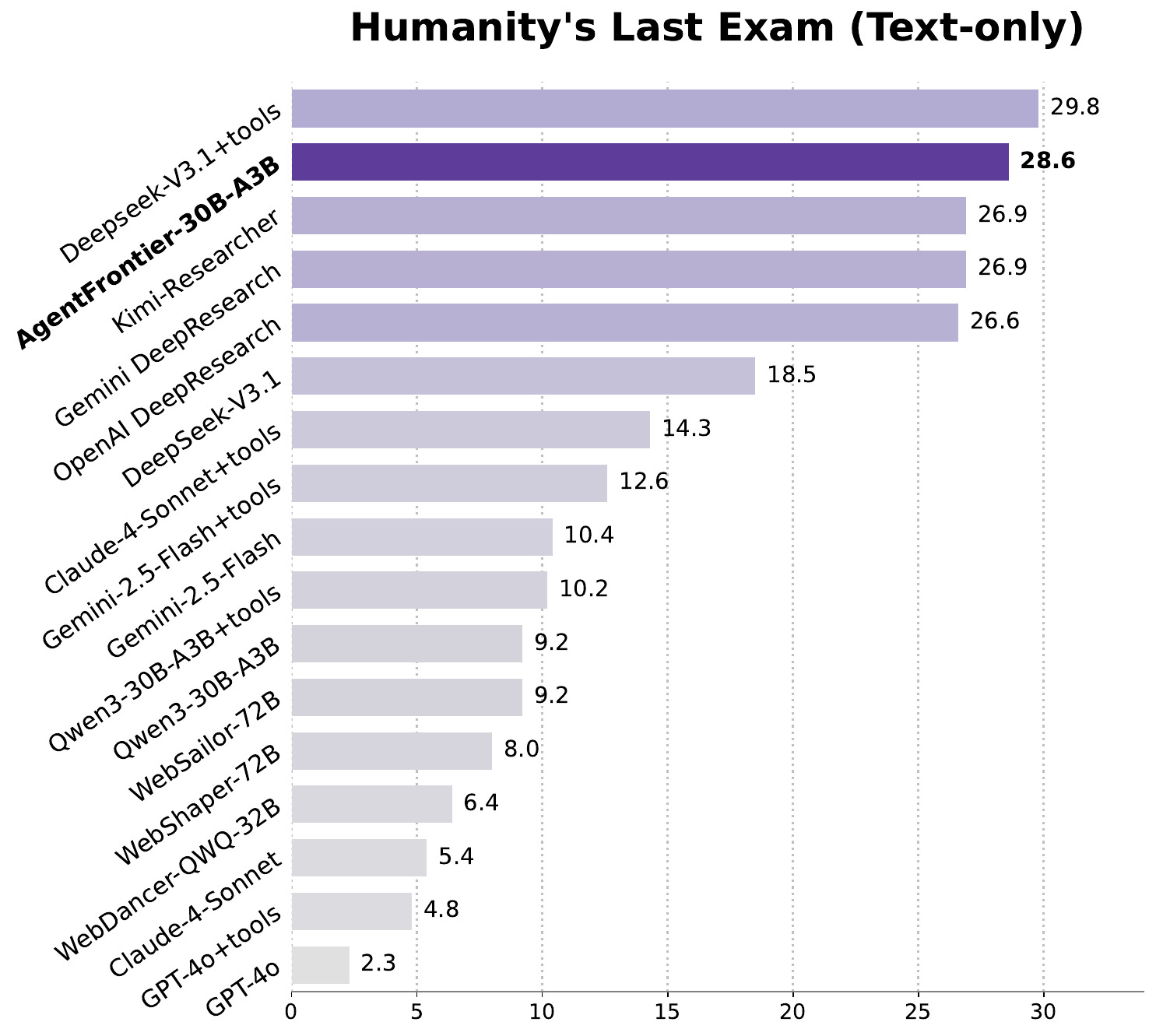}
    \caption{Humanity's Last Exam (Text-only) Results.}
    \label{fig:hle_res}
\end{subfigure}
\hfill 
\begin{subfigure}[b]{0.48\textwidth}
    \centering
    \includegraphics[width=\linewidth]{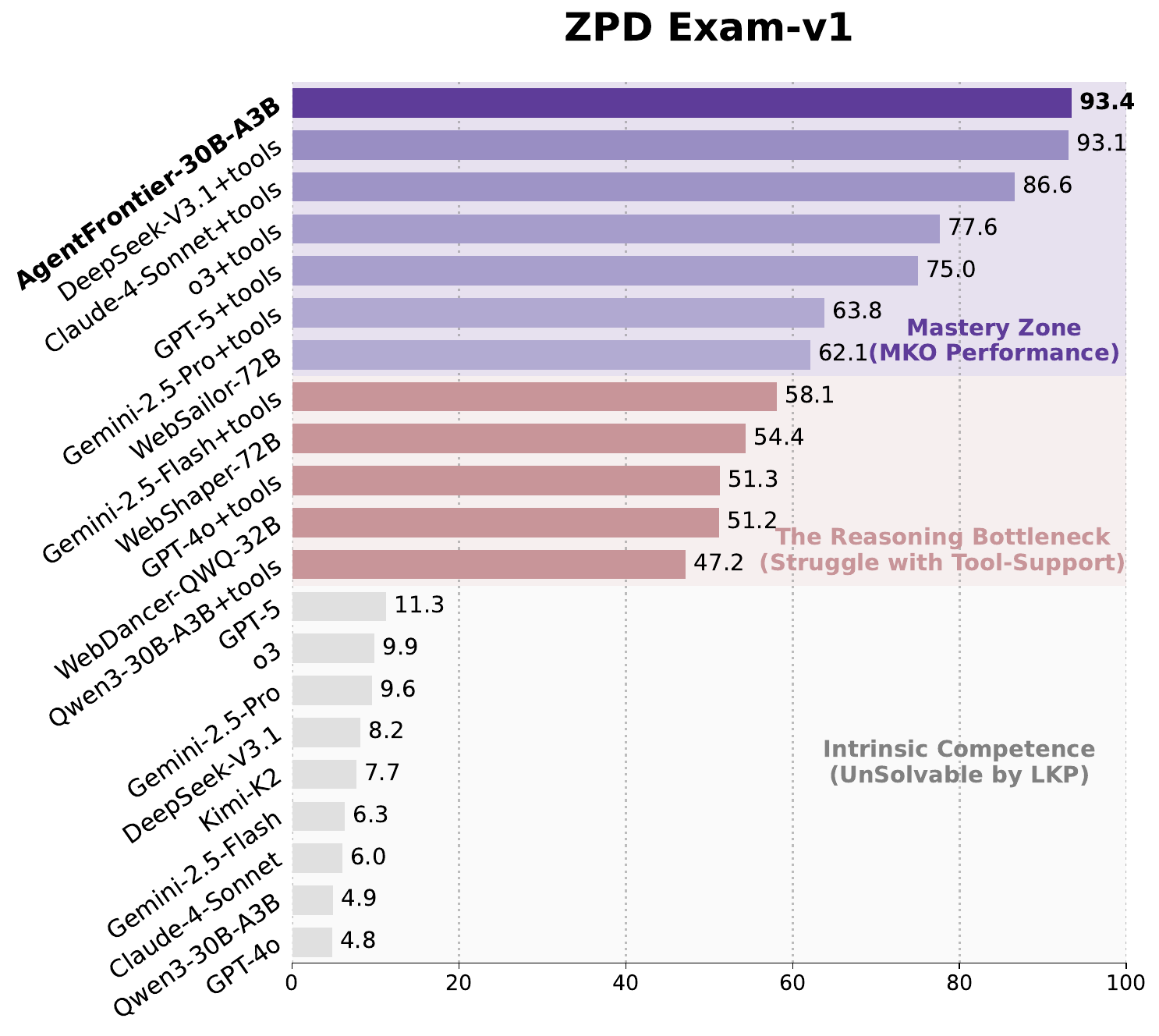}
    \caption{ZPD Exam-v1 Results.}
    \label{fig:zpd_res}
\end{subfigure}
\caption{Performance of LLM agents on the text-only HLE text-only set and ZPD Exam-v1.}
\label{fig:main_results}
\vspace{-15pt}
\end{figure*}

\newpage

\newpage
\section{Introduction}
\label{sec:intro}

Large language models (LLMs) have demonstrated impressive proficiency on various fundamental reasoning tasks~\citep{gpqa, mmlupro, tian2024scicode}. However, they still struggle with the scenarios demanding in-depth, cross-domain, and integrative reasoning~\citep{mialon2023gaia, bc_en, hle}. This gap presents a critical impediment in the pursuit of artificial general intelligence (AGI). Achieving such a leap requires LLMs to move beyond internal knowledge toward agentic behavior, encompassing tool using~\citep{qin2024toolllm}, self-reflection~\citep{shinn2023reflexion}, iterative planning, and multi-step reasoning. The development of such abilities is slowed by the deficit in existing training corpora, which provide little systematic support for cultivating these agentic skills in a unified manner~\citep{shi2025taskcraft}. Besides the scarcity of high-quality training resources, progress is further constrained by the saturation of existing benchmarks and the absence of scalable methods for synthesizing challenging data that reflects the frontiers of human knowledge. While expert-crafted evaluations such as \textit{Humanity's Last Exam}~\citep{hle} offer invaluable benchmarks, their prohibitive cost and lack of scalability underscore the urgent need for automated, frontier-level data synthesis pipelines.

Recent datasets have significantly enhanced LLMs' single-step reasoning~\citep{liu2025synlogic}, but they seldom target the deeper challenge of \textbf{knowledge fusion}~\citep{wan2024knowledge}: integrating and transforming information across diverse sources. While retrieval-augmented generation (RAG)~\citep{lewis2020retrieval} excels when the answer can be grounded in a single document, its performance degrades on tasks requiring reasoning across heterogeneous information. This deficiency traces back to the dominant data-synthesis paradigms, which fall into two broad categories: query-centric methods~\citep{yan2025recitation} that generate variations of existing question–answer (QA) pairs, and document-centric methods~\citep{fan2025megascience, yuan2025naturalreasoning} that derive document-grounded QA pairs from the corpus. Both approaches primarily assess localized comprehension, akin to examining a student on individual textbook chapter rather than their ability to synthesize insights across an entire curriculum. In contrast, complex real-world tasks such as academic research, legal analysis, or engineering design demand multi-document synthesis and cross-domain knowledge fusion. Human experts rarely treat information in isolation; instead, they connect, contrast, and integrate it to derive in-depth insights, which is the intrinsic essence of \textbf{deep research}~\citep{dr, google_dr}. Cultivating this synthetic reasoning capacity in LLMs is paramount for advancing toward higher forms of intelligence.

\begin{wrapfigure}{r}{0.4\textwidth} 
    \vspace{-30pt}
    \centering
    \includegraphics[width=\linewidth]{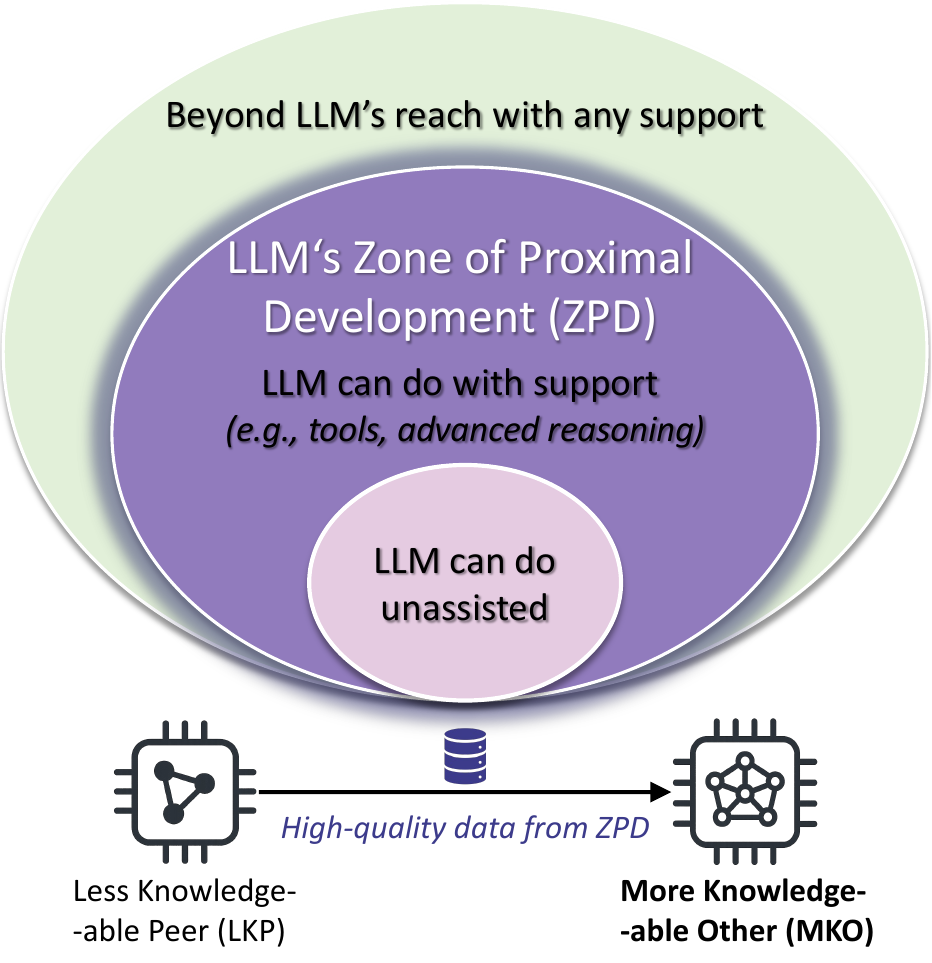}
    \caption{High-quality data situated in an LLM's ZPD acts as a catalyst, transforming it from a LKP into a MKO.}
    \label{fig:ZPD}
    \vspace{-4pt}
\end{wrapfigure}

The central challenge of data synthesis is not merely generating difficult tasks, but calibrating their difficulty to the precise frontier of a model's competence: complex enough to exceed the boundary of the model's intrinsic competence, yet solvable with appropriate support. Existing approaches typically rely on coarse-grained difficulty annotations~\citep{su-etal-2025-nemotron} or heuristically stacked constraints~\citep{patel2025get}, without a precise mechanism for targeting this frontier. In practice, self-generated approaches tend to yield data that remain within the model's own expressive ceiling, making difficulty escalation noisy and unscalable. To address this, we draw inspiration from the educational psychology concept of the \textit{\textbf{Zone of Proximal Development}} (ZPD)~\citep{vygotsky1978mind, mcleod2012zone}, which defines the cognitive space where a learner cannot solve tasks independently but can succeed with guidance. We operationalize this by defining two personas: the \textbf{Less Knowledgeable Peer} (LKP), a base LLM without tools, and the \textbf{More Knowledgeable Other} (MKO), a superior tool-augmented agent with advanced reasoning. Training data unsolvable by the LKP but solvable by the MKO is by definition situated at the model's capability frontier, offering maximally informative supervision. As the model learns, its ZPD advances, enabling a continuously adaptive curriculum.

Collectively, we instantiate this principle in the \textbf{\myname~Engine}, a novel data synthesis framework designed to automatically generate complex-reasoning data within LLM's ZPD. The engine operates through a process of adversarial calibration, dynamically probing the capability frontier of the LLMs. It systematically constructs multidisciplinary QA that necessitate knowledge fusion across multiple web documents, moving beyond simple fact retrieval. Knowledge-intensive data tasks solvable by the LKP are retained for continued pre-training (CPT), while tasks solvable only by the MKO are marked as frontier-level data for post-training. This dual-pipeline design yields a continuous stream of adaptive, high-quality training data, establishing a virtuous cycle of capability growth.

Our contributions are threefold:
\begin{enumerate}
    \item We present \textbf{\myname~Engine}, a scalable data synthesis framework founded on the theory of \textit{Zone of Proximal Development} (ZPD). By integrating agentic refinement and LKP–MKO adversarial calibration, our engine create both knowledge-intensive and frontier-level reasoning data.
    \item We establish \textbf{ZPD Exam}, an automated benchmark designed to probe the ZPD of LLMs. It assesses advanced capabilities such as tool using and in-depth reasoning by complex multidisciplinary questions that require cross-document knowledge fusion and deep research.
    \item We build \textbf{\myname-30B-A3B} by further training Qwen3-30B-A3B-Thing-2507. The model was continually pre-trained on 50 billion tokens of knowledge-intensive data and then post-trained on 12,000 frontier-level QA trajectories synthesized by our engine, achieving 28.6\% on HLE, as well as state-of-the-art performance on ZPD Exam-v1, R-Bench-T and xBench-ScienceQA.
\end{enumerate}

\section{\myname~Data Engine}
\label{sec:data_engine}

\begin{figure}[h]
    \centering
    \includegraphics[width=\linewidth]{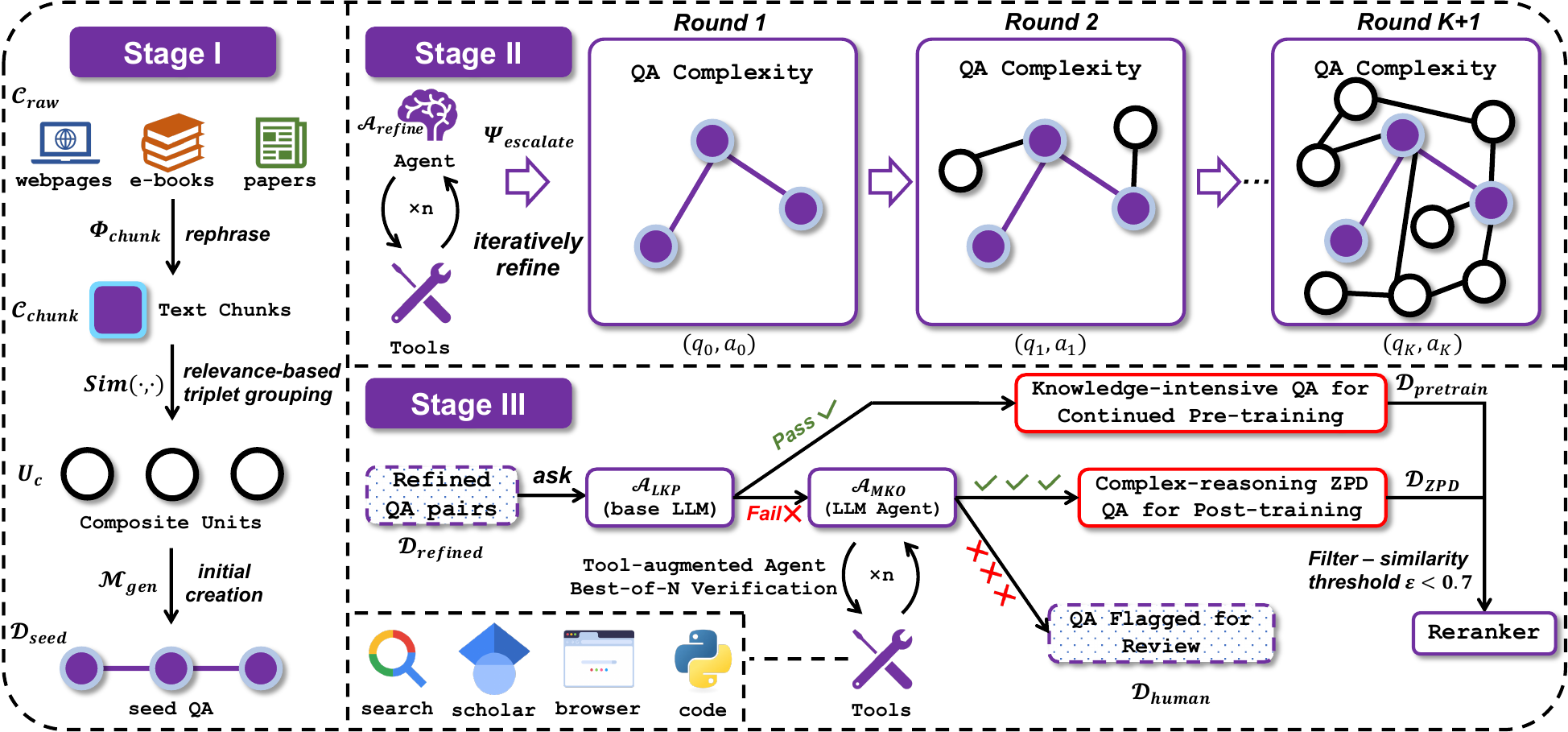}
    \caption{The three-stage synthesis pipeline of the \myname~Engine. It begins by creating multi-source seed questions, then iteratively escalates their complexity using a tool-augmented agent, and finally filters through our ZPD-based calibration mechanism to isolate high-value training data.}
    \label{fig:data_engine}
\end{figure}

\textbf{\myname~Engine} addresses the critical need for training data that fosters knowledge fusion and complex reasoning, which operationalizes the theoretical framework of the \textit{Zone of Proximal Development} to generate challenging tasks that reside at the frontier of a LLM's capabilities. Instead of passively curating existing information, the engine is designed to actively forge complexity through a three-stage agentic synthesis pipeline. This process aims to evolve LLMs from knowledge retrievers into sophisticated reasoning agents. The entire workflow, depicted in Figure~\ref{fig:data_engine}, transforms a raw document corpus $\mathcal{C}_{\text{raw}}$ into a calibrated, high-value dataset $\mathcal{D}_{\text{ZPD}}$. The detailed procedure is presented in Algorithm~\ref{alg:data_engine_revised}.

\subsection{Stage I: Seed Question Generation for Knowledge Fusionn}

The pipeline begins with a diverse, multi-disciplinary corpus $\mathcal{C}_{\text{raw}}$ of one million public documents. We first employ a powerful LLM, Qwen3-235B-A22B~\citep{yang2025qwen3}, as a chunking function $\Phi_{\text{chunk}}$ to preprocess the corpus. This function cleans artifacts (e.g., HTML tags) and condenses long texts into information-dense chunks $\mathcal{C}_{\text{chunk}}$, such that $\mathcal{C}_{\text{chunk}} = \bigcup_{d \in \mathcal{C}_{\text{raw}}} \Phi_{\text{chunk}}(d)$.

To generate tasks that inherently demand knowledge fusion, we synthesize questions from \textbf{composite units}---groups of thematically related chunks. To overcome the computational infeasibility of a combinatorial search, we adopt an efficient, retrieval-based approach. We first build a vector index over $\mathcal{C}_{\text{chunk}}$ and, for each chunk $c_i$, retrieve its $k_{\text{nn}}$ nearest neighbors. Within this local neighborhood, we search for triplets $(c_i, c_j, c_k)$ that exhibit high thematic coherence, formally defined as $\text{Sim}(c_x, c_y) > \tau_{\text{theme}}$ for all distinct pairs, where $\text{Sim}(\cdot, \cdot)$ is a semantic similarity function.

These composite units are then fed to a generator model, $\mathcal{M}_{\text{gen}}$, to synthesize initial question-answer pairs. This process yields a seed dataset that serves as the foundation for complexity escalation: $\mathcal{D}_{\text{seed}} = \{ (q_0, a_0) = \mathcal{M}_{\text{gen}}(U_c) \mid U_c \text{ is a composite unit} \}$.

\subsection{Stage II: Escalating Complexity through Agentic Refinement}

The core of our engine is an iterative refinement loop driven by a refinement agent $\mathcal{A}_{\text{refine}}$ with a tool suite $\mathcal{T} = \{T_{\text{search}}, T_{\text{scholar}}, T_{\text{browser}}, T_{\text{code}}\}$. For a QA pair $(q_k, a_k)$ at iteration $k$, the agent applies an escalation operator $\Psi_{\text{escalate}}$ to generate a more sophisticated pair $(q_{k+1}, a_{k+1}) = \Psi_{\text{escalate}}(q_k, a_k, \mathcal{A}_{\text{refine}})$. This operator enriches the QA along four dimensions:
\begin{itemize}
    \item \textbf{Knowledge Expansion:} It actively queries external sources to retrieve and weave in relevant background knowledge, broadening the informational scope of the question.
    \item \textbf{Conceptual Abstraction:} It conducts in-depth analysis of the core concepts within the provided materials, abstracting higher-level principles or identifying subtle relationships.
    \item \textbf{Factual Grounding:} It performs multi-source cross-validation and targeted augmentation to enhance the factual accuracy and depth of the content.
    \item \textbf{Computational Formulation:} It leverages the Python execution to craft QA that require quantitative calculation or logical simulation, assessing reasoning and computational skills.
\end{itemize}

\begin{figure}[h]
    \centering
    \includegraphics[width=\linewidth]{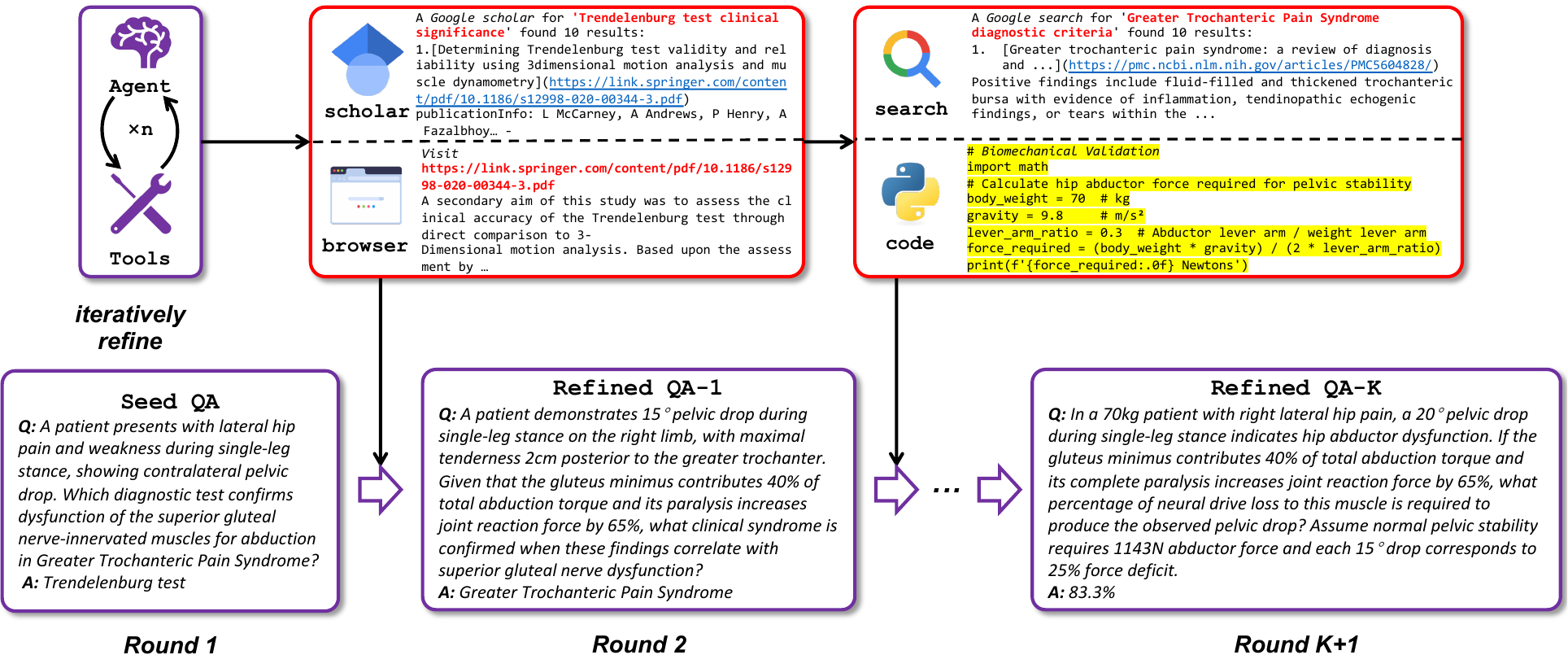}
    \caption{An overview of our iterative refinement process. We start with a biomedical seed QA, which is then refined into a complex diagnostic reasoning problem by synthesizing knowledge from academic literature. Finally, this problem is evolved into a practical computational challenge grounded in a real-world application, a process involving web search and programmatic validation.}
    \label{fig:refined_qa}
\end{figure}

This self-bootstrapping process creates a virtuous cycle, where the output of one iteration becomes the input for the next, building increasingly more intricate reasoning paths. Figure~\ref{fig:refined_qa} illustrates an example where a question is progressively refined by interleaving web search with numerical computation. After $K$ iterations, this stage produces a dataset of highly complex QA pairs, $\mathcal{D}_{\text{refined}}$.

\subsection{Stage III: ZPD-based Filtering and Calibration}

Not all synthesized QA pairs are equally valuable for training. To isolate tasks that reside precisely within an LLM's ZPD, we introduce a rigorous calibration mechanism based on our \textbf{LKP-MKO} framework. We instantiate a \textbf{Less Knowledgeable Peer} ($\mathcal{A}_{\text{LKP}}$) with the base LLM and a \textbf{More Knowledgeable Other} ($\mathcal{A}_{\text{MKO}}$) with the powerful, tool-augmented agent.

For each candidate pair $(q, a) \in \mathcal{D}_{\text{refined}}$, we first assess its difficulty. Let $\text{IsSolvableBy}(\mathcal{A}, q, a) \in \{0, 1\}$ be a binary function, implemented by an automated judge (GPT-4o~\citep{gpt4o}), which returns 1 if agent $\mathcal{A}$ correctly answers $q$. (a) If $\text{IsSolvableBy}(\mathcal{A}_{\text{LKP}}, q, a) = 1$, the pair is deemed too simple and is allocated to a general knowledge dataset $\mathcal{D}_{\text{pretrain}}$ for continued pre-training. (b) If $\text{IsSolvableBy}(\mathcal{A}_{\text{LKP}}, q, a) = 0$, the pair is challenging and passed to the MKO for further evaluation.

To stratify the challenging data, $\mathcal{A}_{\text{MKO}}$ performs Best-of-N (BoN) verification with $N=3$, generating $N$ independent solutions $\{s_1, \dots, s_N\}$. The data is then partitioned based on the outcome:
\begin{itemize}
    \item \textbf{Verified for Post-Training ($\mathcal{D}_{\text{ZPD}}$):} If the MKO finds at least one correct solution (i.e., $\sum_{i=1}^{N} \text{IsCorrect}(s_i, a) \ge 1$), the pair is considered to be within the model's ZPD—challenging yet learnable. These verified pairs form our final training set.
    \item \textbf{Flagged for Human Review ($\mathcal{D}_{\text{human}}$):} If the MKO fails in all $N$ attempts (i.e., $\sum_{i=1}^{N} \text{IsCorrect}(s_i, a) = 0$), the pair is either flawed or exceptionally difficult and is routed to human experts for analysis.
\end{itemize}

Finally, to ensure dataset diversity, we apply a semantic redundancy filter. A newly generated pair $(q', a')$ is discarded if its question $q'$ is too similar to any question already in $\mathcal{D}_{\text{ZPD}}$. Specifically, we discard $(q', a')$ if $\max_{(q, a) \in \mathcal{D}_{\text{ZPD}}} \text{Sim}(q', q) \ge \epsilon$, where $\text{Sim}(\cdot, \cdot)$ is measured by a reranker model~\citep{qwen3embedding} and the threshold $\epsilon$ is set to 0.7.

Through this three-stage pipeline, the \myname~Engine provides a scalable method for generating complex reasoning data, continuously pushing the boundaries of LLM capabilities.
\section{ZPD Exam: A Self-Evolving Benchmark for LLM Agents}
\label{sec:wwe}

Evaluating rapidly advancing LLMs requires benchmarks that co-evolve with their capabilities. While expert-crafted exams like Humanity's Last Exam~\citep{hle} probe the frontier of human knowledge, their static nature and prohibitive creation costs hinder scalable and continuous assessment. We introduce the \textbf{ZPD Exam}, an automated and continuously evolving benchmark designed to assess the deep research capabilities of advanced LLM agents.

\subsection{Benchmark Construction: From Frontier Knowledge to Agentic Research}
The ZPD Exam is designed to simulate scientific discovery by generating tasks that are intractable using only parametric knowledge, thus compelling models to function as research agents. The benchmark is constructed using our \myname~Engine (Section~\ref{sec:data_engine}), specifically configured to generate novel, multi-disciplinary questions. Crucially, this benchmark corpus is strictly disjoint from the corpus used to construct our training data, ensuring a fair and uncontaminated evaluation.

\textbf{Grounding in the Knowledge Frontier.} We ground this exam in the knowledge frontier by curating a corpus of 30,000 recent scientific papers published between 2023 and 2025, spanning multi-disciplinary domains such as mathematics, computer science, and physics. This ensures that success demands genuine, on-the-fly reasoning and information synthesis, not merely knowledge retrieval.

\textbf{Calibrating Tasks to the LLM's ZPD.} From our initial corpus, the \myname~Engine generates candidate questions, which are then subjected to a strict adversarial filter to align with the ZPD of a baseline model. To be included in ZPD Exam-v1, a problem must satisfy a dual constraint: it must be unsolvable by the baseline model in three unaided attempts, yet consistently solvable by the same model across three attempts when granted access to tools. This process isolates problems that are difficult but solvable with assistance, defining the empirical boundary of the model's ZPD.

\begin{figure}[h]
    \centering
    \includegraphics[width=\linewidth]{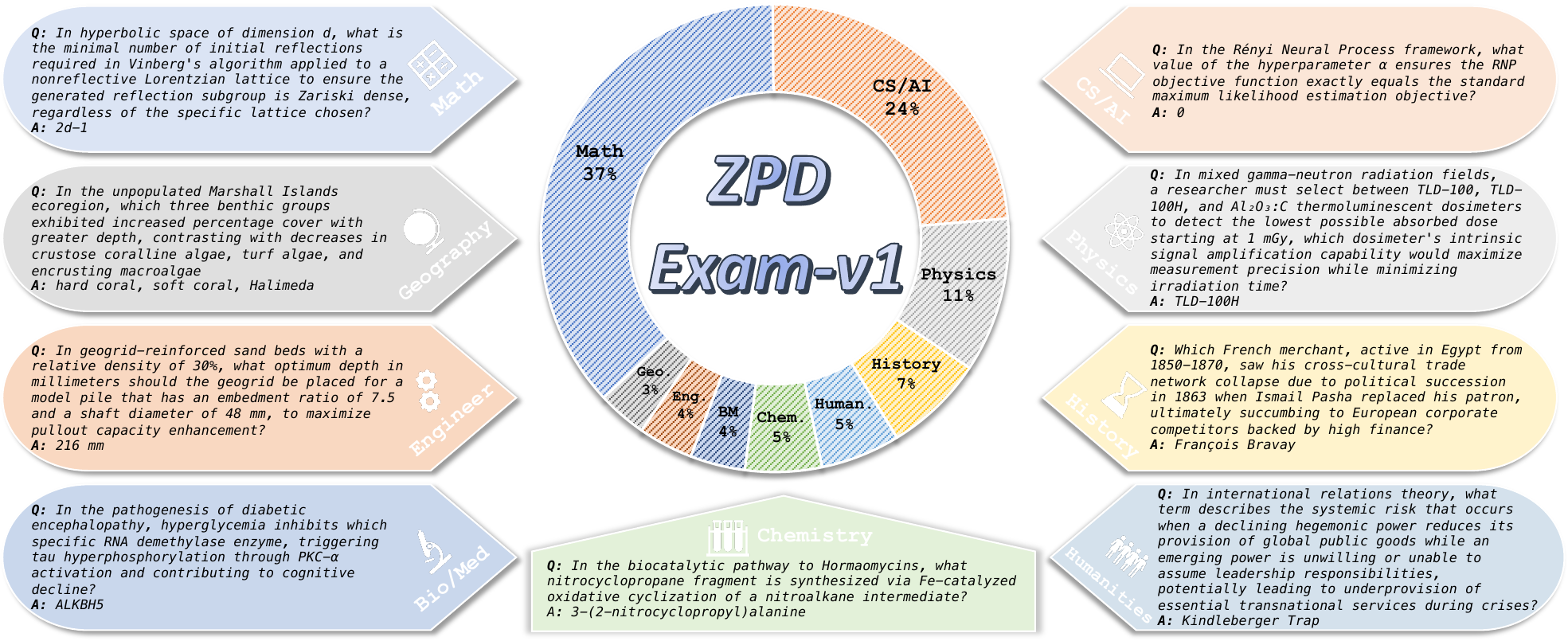}
    \caption{The ZPD Exam-v1 consists of 1024 questions categorized into 9 disciplines: Mathematics, Computer Science / Artificial Intelligence, Physics, History, Humanities, Chemistry, Biology / Medicine, Engineering, and Geography.}
    \label{fig:wwe}
\end{figure}

This automated pipeline enables a flywheel-like iterative process: as models improve, the ZPD exam can be regenerated to target the new frontier, making it a \textbf{living benchmark} resistant to saturation. After multiple rounds of validation and deduplication, ZPD Exam-v1 was constructed by sampling 1,024 public questions and a corresponding private set. All questions are open-ended short-answer format, facilitating automated grading. The benchmark composition is detailed in Figure~\ref{fig:wwe}.

\subsection{ZPD Exam: A Diagnostic Benchmark for Agentic Reasoning}
The ZPD Exam proposes a new evaluative framework, shifting the focus from an LLM's static parametric knowledge~\citep{mmlu} to its dynamic capacity for knowledge discovery, which functions as an "open-book" examination where agent must first author the "book" through active exploration and tool use. This design philosophy deliberately situates the challenges within the ZPD for current LLMs, a calibration confirmed by their low initial scores (Figure~\ref{fig:zpd_res}). Our empirical results validate this diagnostic power, revealing a clear stratification of agent performance into three distinct zones.

\textbf{Zone 1: Intrinsic Competence (Score \textless~20).}
This tier establishes the baseline, reflecting the performance of LLMs relying solely on their parametric knowledge (e.g., GPT-5 and Gemini-2.5-Pro without tools). By design, the problems are intractable without external information, confirming that these tasks lie outside the models' unaided capabilities. This zone effectively establishes a baseline, quantifying the limits of intrinsic, "closed-book" reasoning, confirming that any score above this threshold is directly attributable to the agent's ability to leverage external tools support.

\textbf{Zone 2: The Reasoning Bottleneck (Score 20-60).}
This intermediate tier characterizes the ZPD itself, where agents (e.g., GPT-4o with tools, WebShaper-72b) can achieve partial success with assistance but lack mastery. This zone highlights the benchmark's crucial distinction from standard RAG evaluations. While RAG tests comprehension of a given context, agents here falter in the more demanding task of autonomously discovering, structuring, and reasoning over the necessary information. Their failures stem not from tool-level errors but from a higher-order "reasoning bottleneck": a deficit in strategic planning, synthesizing information across multiple tool calls, and adapting their approach. This reveals that access to tools is necessary but insufficient; the primary limiting factor is the agent's meta-cognitive ability to orchestrate these tools effectively.

\textbf{Zone 3: Emergent Mastery (Score \textgreater~60).}
Agents in this top tier (e.g., DeepSeek-V3.1 with tools) demonstrate a qualitative leap in capability. They have transcended the reasoning bottleneck and exhibit robust, multi-step planning and synthesis. Their behavior is analogous to the More Knowledgeable Other, seamlessly integrating tool-based exploration into a coherent reasoning process to solve problems far beyond their intrinsic reach. Achieving this level of performance signifies the emergence of a truly capable agent that can autonomously navigate complex problem spaces.

In summary, the ZPD Exam serves not merely as a leaderboard but as a powerful diagnostic instrument. Its tiered results provide a fine-grained analysis of an agent's developmental stage—from what it knows (intrinsic), to what it can learn to do with support (ZPD), to what it has mastered. This allows us to pinpoint critical reasoning faculties that require improvement, thereby charting a clear path toward more autonomous and capable AI agents.
\section{Experiments}

\subsection{Experimental Setup}

\paragraph{Training Data Synthesis} We synthesize training trajectories using a tool-augmented agent, following the iterative tool-calling and summarization paradigm from WebResearcher~\citep{qiao2025webresearcher}. Each trajectory is generated through a multi-round process adhering to the ReAct~\citep{yao2023react}, comprising a sequence of round-wise reasoning reports and observations after the corresponding tool calls. In each round, the model generates a reasoning report that summarizes accumulated evidence, analyzes progress towards the research question, and specifies the next action—either invoking a new tool or outputting a final answer.

\paragraph{Rejection Sampling Fine-Tuning} Formally, given a research question $q^{(i)}$, the model generates the reasoning report $r_j^{(i)}$ at round $j$ conditioned on the previous report–observation pair $\{r_{j-1}^{(i)}, o_{j-1}^{(i)}\}$, with initialization $r_0^{(i)} = o_0^{(i)} = \emptyset$.  For a collection of $K$ accepted trajectories, where trajectory $i$ has $L_i$ rounds, the objective reduces to
supervised learning that maximizes the conditional log-likelihood:
\begin{equation}
\mathcal{L}_{\text{RFT}}(\theta) = - \ \sum_{i=1}^{K} \sum_{j=1}^{L_i} \log p_\theta \Big( r_j^{(i)} \,\Big|\, q^{(i)}, r_{j-1}^{(i)}, o_{j-1}^{(i)} \Big),
\end{equation}
where $\theta$ denotes the model parameters. The loss computed is exclusively on the reasoning report tokens; tool observations are included in the context but excluded from backpropagation.

\paragraph{Models and Benchmarks} We apply RFT to the Qwen3 family of models~\citep{yang2025qwen3}, including both dense (Qwen3-8B, Qwen3-32B) and mixture-of-experts (Qwen3-30B-A3B-Thinking-2507) variants. We evaluate performance on four challenging benchmarks designed to probe high-level reasoning across diverse disciplines:
\begin{itemize}
    \item \textbf{HLE}~\citep{hle} - Humanity's Last Exam is an expert-curated benchmark of 2,500 highly challenging questions spanning a wide range of disciplines, designed to assess frontier-level academic competence. We use the 2,154 text-only questions.
    \item \textbf{ZPD Exam} - Our newly proposed multidisciplinary benchmark designed to probe the LLM's zone of proximal development. We use the 1,024 questions from its first version.
    \item \textbf{R-Bench}~\citep{guorbench} - A graduate-level, multidisciplinary benchmark designed to comprehensively assess the complex reasoning capabilities of LLMs. We used its English text-only version. After excluding one question for potential ambiguity, our evaluation set consists of 1,093 multiple-choice questions.
    \item \textbf{xBench-ScienceQA}~\citep{xbench} - A curated set of 100 Chinese QA items from the xBench suite, designed to evaluate foundational scientific knowledge.
\end{itemize}

\paragraph{Baselines}

We evaluate our proposed \myname~dataset by comparing it with three well-established, multidisciplinary public datasets for agent fine-tuning:
\begin{itemize}
    \item \textbf{TaskCraft}~\citep{shi2025taskcraft} - The TaskCraft dataset facilitates the fine-tuning of agent models by programmatically generating agentic tasks at scale. These tasks are characterized by their inclusion of multiple tools, tiered difficulty levels, and verifiable execution trajectories.
    \item \textbf{MegaScience}~\citep{fan2025megascience} - The MegaScience dataset is constructed by integrating high-quality subsets from multiple open-source scientific datasets to ensure sample abundance and high fidelity. The majority of its questions are sourced from university textbooks.
    \item \textbf{MiroVerse}~\citep{miromind2024opendata} - MiroVerse is an open-source, large-scale dataset for AI agents, covering diverse tasks such as multi-hop question answering, web navigation, and scientific reasoning. We use the SFT data from its v0.1 release.
\end{itemize}

For each dataset, we first curate 12,000 high-quality trajectories via rejection sampling, retaining only instances where the model's final answer perfectly matches the ground truth. As shown in Table~\ref{tab:data_stats}, our \myname~dataset exhibits a more balanced and diverse tool-use distribution compared to the baselines, with substantial usage across scholar, browser, and code tools. This reflects its focus on complex, knowledge-intensive problem-solving. To ensure a fair comparison, we normalize the training data volume to 25,600 rounds for each dataset, with each round capped at 40,960 tokens, and train for 3 epochs.

\begin{table}[htbp]
\centering
\caption{Statistics of the training datasets. "Avg. Rounds" and "Avg. Calls" are computed per trajectory.}
\label{tab:data_stats}
\begin{tabular}{l c cccc}
\toprule

\multirow{2}{*}{\textbf{Dataset}} & \multirow{2}{*}{\textbf{Avg. Rounds}} & \multicolumn{4}{c}{\textbf{Avg. Calls}} \\
\cmidrule(lr){3-6}
& & \textbf{Search} & \textbf{Scholar} & \textbf{Browser} & \textbf{Code} \\
\midrule
TaskCraft   & 3.38 & 1.04 & 0.14 & 1.19 & 0.01 \\
MegaScience & 2.68 & 0.26 & 0.56 & 0.49 & 0.37 \\
MiroVerse   & 2.18 & 0.12 & 0.04 & 0.09 & 0.93 \\
\myname     & 3.32 & 0.32 & 0.66 & 0.82 & 0.52 \\
\bottomrule
\end{tabular}
\end{table}

\paragraph{Hyper-parameters and Metric} For all generation tasks, we use nucleus sampling with a \textbf{temperature} of 0.6 and a \textbf{top-p} of 0.95. To evaluate the correctness of the final answers, we employ an \textbf{LLM-as-a-Judge}. Specifically, we use o3-mini~\citep{o3} as the judge, guided by the official strict evaluation prompt from HLE~\citep{hle}, to assess the correctness of model responses against the ground truth.

\subsection{Main Results} 

\begin{figure}[t]
    \centering
    \begin{subfigure}[b]{0.24\textwidth}
        \centering
        \includegraphics[width=\textwidth]{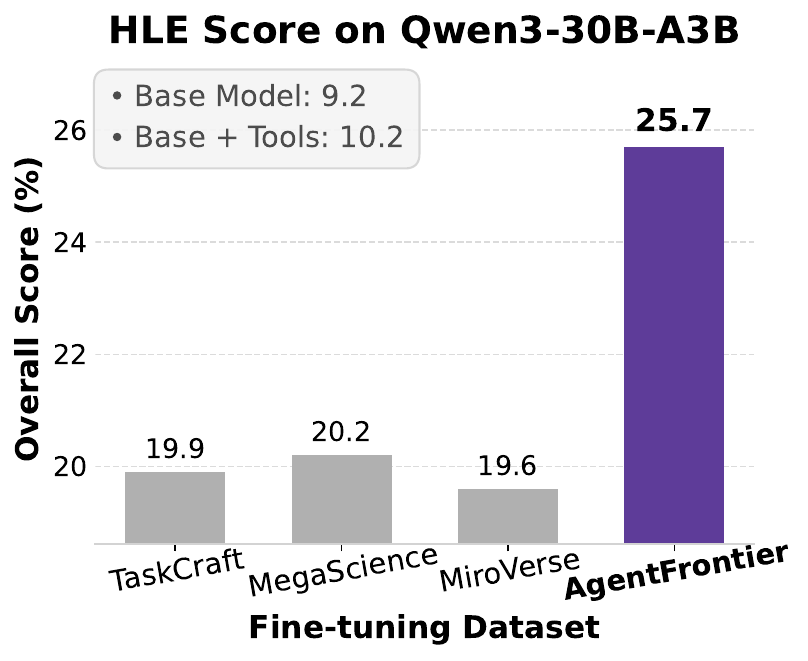}
    \end{subfigure}
    \hfill
    \begin{subfigure}[b]{0.24\textwidth}
        \centering
        \includegraphics[width=\textwidth]{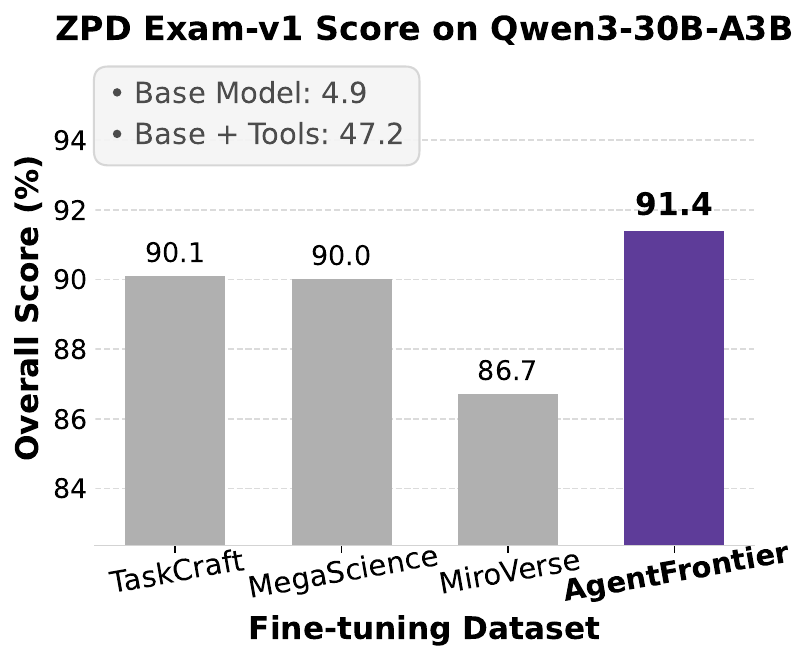}
    \end{subfigure}
    \hfill
    \begin{subfigure}[b]{0.24\textwidth}
        \centering
        \includegraphics[width=\textwidth]{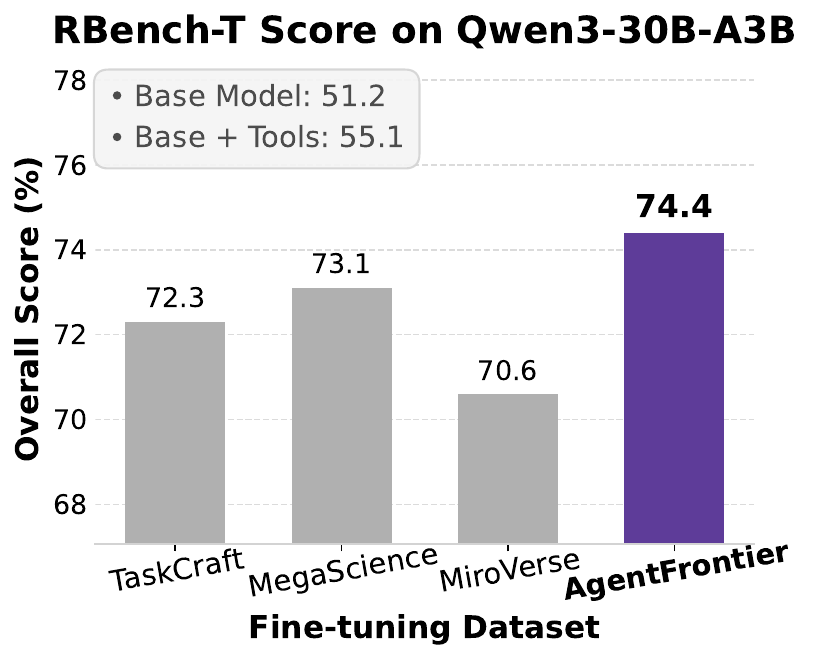}
    \end{subfigure}
    \hfill 
    \begin{subfigure}[b]{0.24\textwidth}
        \centering
        \includegraphics[width=\textwidth]{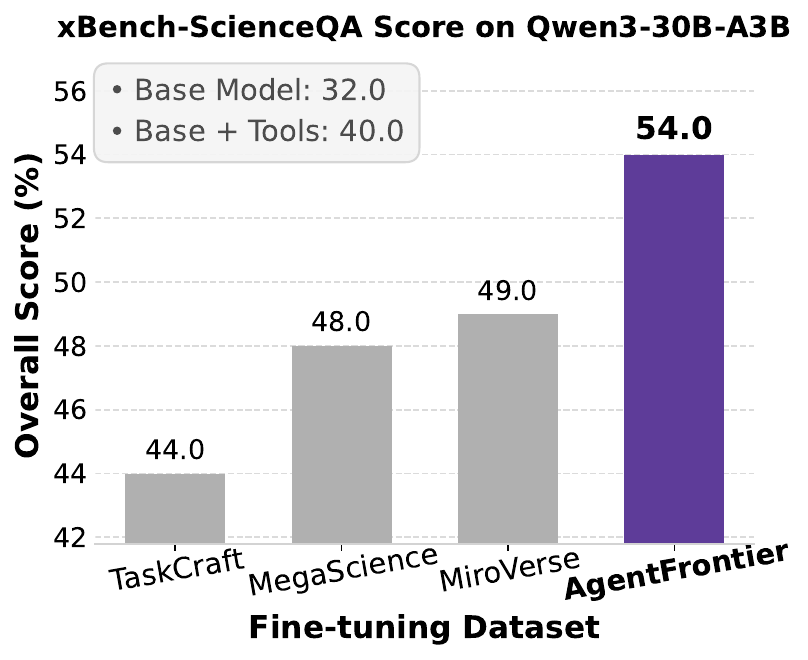}
    \end{subfigure}

    \begin{subfigure}[b]{0.24\textwidth}
        \centering
        \includegraphics[width=\textwidth]{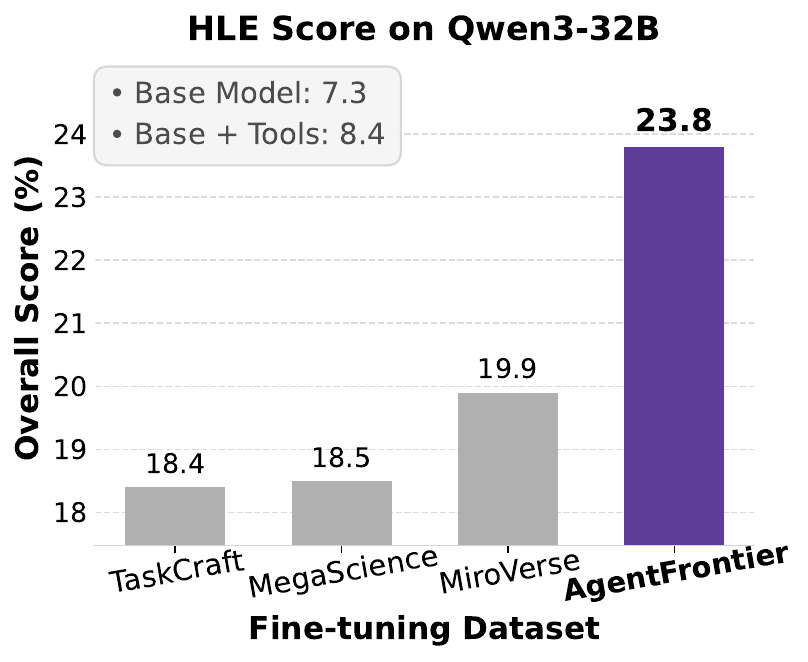}
    \end{subfigure}
    \hfill
    \begin{subfigure}[b]{0.24\textwidth}
        \centering
        \includegraphics[width=\textwidth]{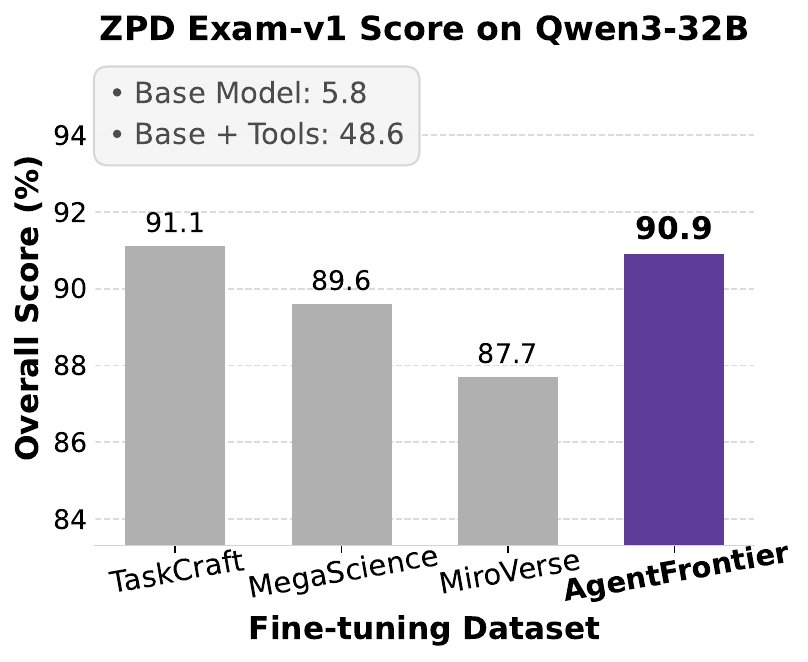}
    \end{subfigure}
    \hfill
    \begin{subfigure}[b]{0.24\textwidth}
        \centering
        \includegraphics[width=\textwidth]{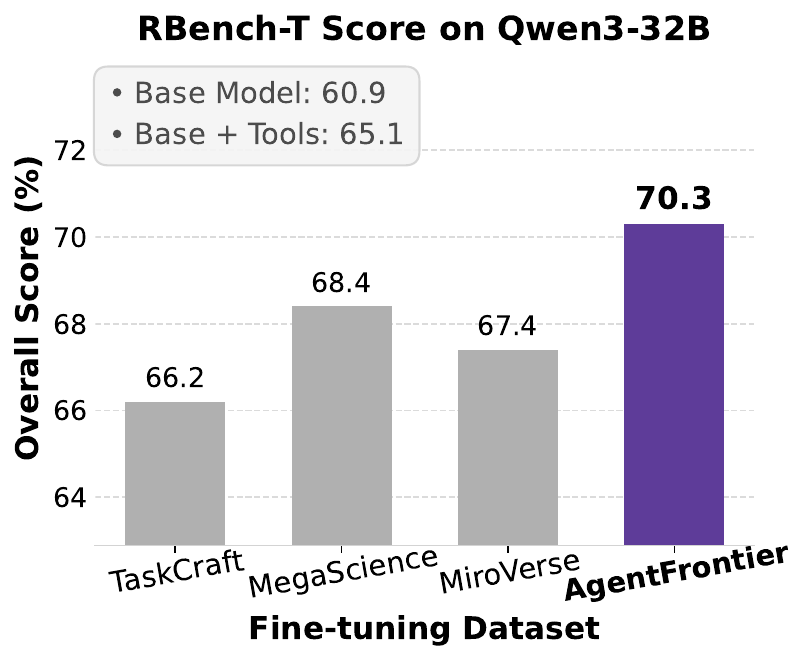}
    \end{subfigure}
    \hfill 
    \begin{subfigure}[b]{0.24\textwidth}
        \centering
        \includegraphics[width=\textwidth]{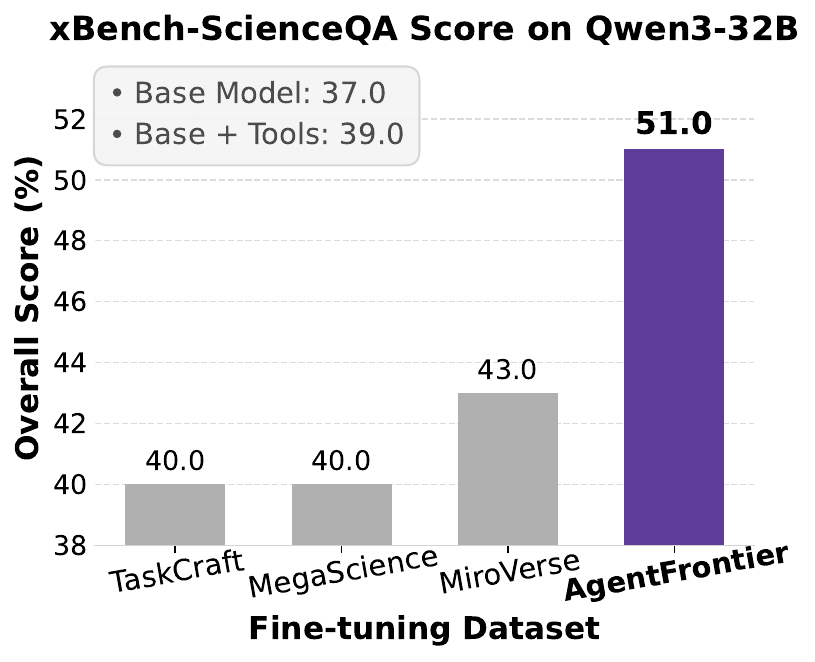}
    \end{subfigure}

    \begin{subfigure}[b]{0.24\textwidth}
        \centering
        \includegraphics[width=\textwidth]{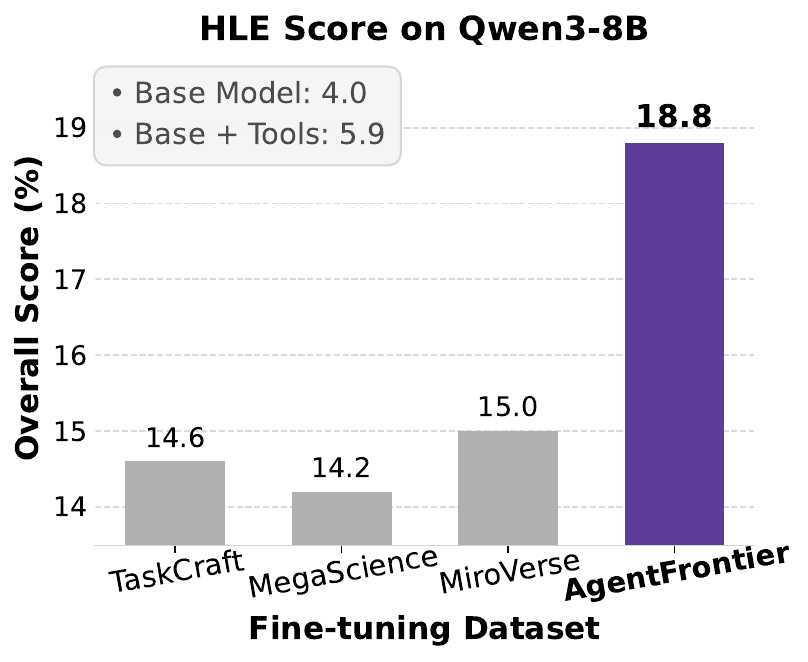}
    \end{subfigure}
    \hfill
    \begin{subfigure}[b]{0.24\textwidth}
        \centering
        \includegraphics[width=\textwidth]{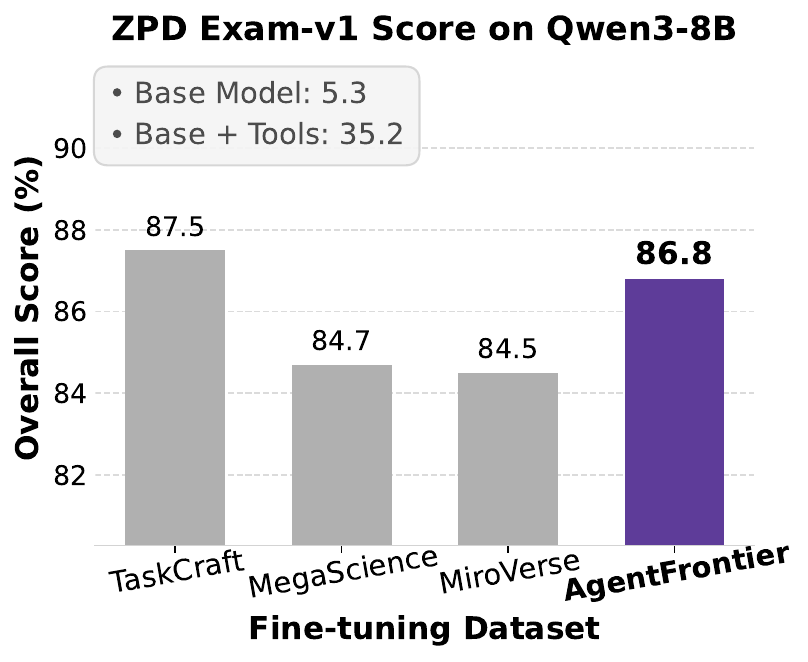}
    \end{subfigure}
    \hfill
    \begin{subfigure}[b]{0.24\textwidth}
        \centering
        \includegraphics[width=\textwidth]{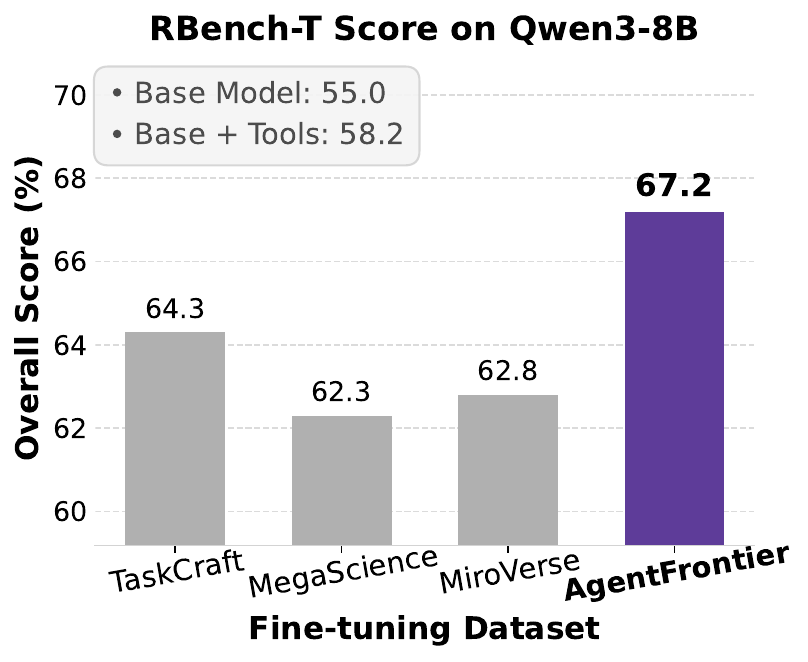}
    \end{subfigure}
    \hfill 
    \begin{subfigure}[b]{0.24\textwidth}
        \centering
        \includegraphics[width=\textwidth]{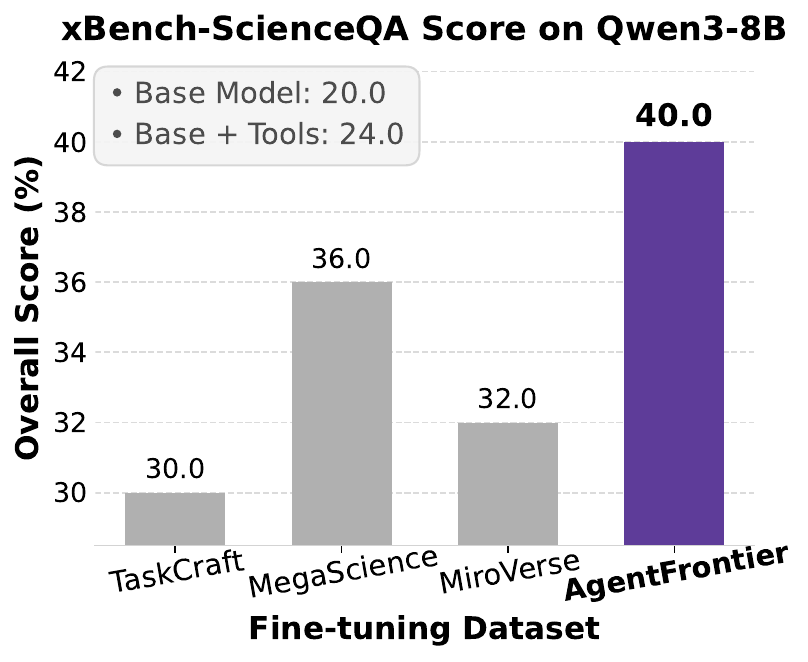}
    \end{subfigure}

    \caption{Impact of fine-tuning datasets on Qwen3 series models' performance across 4 benchmarks.}
    \label{fig:all_benchmarks_turbo}
\end{figure}

\paragraph{Overall Performance Across Benchmarks}
As illustrated in Figure~\ref{fig:all_benchmarks_turbo}, when fine-tuning the Qwen3-30B-A3B model, models trained on \myname~consistently achieve state-of-the-art performance, decisively outperforming all other training datasets across every benchmark evaluated. In contrast, the performance of competing datasets such as TaskCraft, MegaScience, and MiroVerse is inconsistent; while each may show strength on a particular benchmark, none demonstrates the robust, cross-domain superiority imparted by \myname. This trend of consistent outperformance holds for other model backbones as well.

\paragraph{Subject-Level Dominance on the HLE Benchmark}

To investigate the source of this performance advantage, we conduct a fine-grained analysis on the particularly demanding Humanity's Last Exam (HLE)~\citep{hle} benchmark, examining results across eight academic disciplines with various model backbones (Table~\ref{tab:main_results}). For both the Qwen3-8B and Qwen3-32B backbones, models trained on \myname~exhibit remarkable breadth, securing the top performance in six and seven out of the eight subjects, respectively. This subject-level dominance translates to a significant lead in overall average scores, with \myname~surpassing the next-best dataset by 3.8 and 3.9 absolute points on the 8B and 32B models, respectively. The advantage becomes even more pronounced with the Qwen3-30B-A3B model, where fine-tuning on \myname~outperforms all competing datasets in every single subject. This comprehensive superiority results in a final average score of 25.67\%, representing a  178\% and 152\% relative improvement over the original base model in settings without and with tool augmentation, respectively. These results indicate that as model capacity increases, the rich, multi-step reasoning trajectories within \myname~become increasingly effective at unlocking expert-level problem-solving capabilities across a wide spectrum of academic fields.

\definecolor{modelcolor1}{rgb}{0.88, 0.92, 0.98} 
\definecolor{modelcolor2}{rgb}{0.9, 0.98, 0.9}   
\definecolor{modelcolor3}{rgb}{0.95, 0.9, 0.98}   
\newcommand{\hlone}[1]{\cellcolor{modelcolor1!150}{\bfseries #1}}
\newcommand{\hltwo}[1]{\cellcolor{modelcolor2!150}{\bfseries #1}}
\newcommand{\hlthree}[1]{\cellcolor{modelcolor3!150}{\bfseries #1}}

\begin{table}[t]
\centering
\caption{Accuracy on the Humanity's Last Exam (full text-only set). Results are reported across major knowledge domains. Each block corresponds to a different Qwen3 backbone. Numbers with a colored background denote the best within each block; \underline{underlined numbers} denote the second best.}
\label{tab:main_results}
\resizebox{\textwidth}{!}{
\begin{tabular}{@{} l @{\hspace{0.2em}} c *{9}{S[table-format=2.2]} @{}}
\toprule
\multirow{2}{*}{\textbf{RFT Dataset}} & \multirow{2}{*}{\textbf{Tools}} & \multicolumn{9}{c}{\textbf{Domain Accuracy on Humanity's Last Exam (\%)}} \\
\cmidrule(l){3-11}
&& {Math} &  {CS/AI} & {Bio./Med.} & {Physics} & {Humanities} & {Chem.} & {Eng.} & {Other} & {Avg.} \\
\midrule

\rowcolor{modelcolor1} \multicolumn{11}{@{}c}{\textit{Backbone: Qwen3-8B}} \\
{--}       & \xmark & 6.46 & 2.65 & 5.88 & 0.99 & 3.63 & 1.00 & 6.45 & 1.61 & 4.00 \\
{--}       & \cmark & 6.26 & 3.54 & 9.05 & 2.48 & 7.25 & 7.00 & 6.45 & 5.14 & 5.94 \\
TaskCraft  & \cmark & 16.21 & \underline{10.62} & 14.93 & \underline{6.44} & \underline{22.80} & \underline{9.00} & \underline{9.68} & 15.43 & 14.58 \\
MegaScience& \cmark & 14.56 & \underline{10.62} & \hlone{18.10} & 5.94 & 21.76 & \underline{9.00} & \hlone{12.90} & 16.57 & 14.21 \\
MiroVerse  & \cmark & \underline{17.33} & \underline{10.62} & 15.38 & 5.94 & 21.24 & 8.00 & 6.45 & \underline{17.71} & \underline{15.00} \\
\hlone{\myname} & \cmark & \hlone{22.46} & \hlone{14.16} & \underline{16.74} & \hlone{10.40} & \hlone{24.35} & \hlone{11.00} & 6.45 & \hlone{19.43} & \hlone{18.80} \\
\midrule

\rowcolor{modelcolor2} \multicolumn{11}{@{}c}{\textit{Backbone: Qwen3-32B}} \\
{--}       & \xmark & 8.72 & 5.75 & 10.41 & 0.50 & 7.77 & 8.00 & 6.45 & 5.14 & 7.34 \\
{--}      & \cmark & 10.97 & 5.31 & 9.05 & 4.95 & 7.25 & 5.00 & 6.45 & 4.57 & 8.36 \\
TaskCraft  & \cmark & 20.72 & 14.16 & \underline{16.74} & 8.91 & 25.39 & \underline{14.00} & \underline{14.52} & 20.57 & 18.43 \\
MegaScience& \cmark & 21.23 & \underline{14.60} & 14.93 & 6.44 & 29.02 & 12.00 & 11.29 & \underline{21.71} & 18.52 \\
MiroVerse  & \cmark & \underline{22.56} & 14.16 & \underline{16.74} & \underline{10.40} & \hltwo{34.72} & 12.00 & 6.45 & 20.57 & \underline{19.92} \\
\hltwo{\myname} & \cmark & \hltwo{28.21} & \hltwo{16.81} & \hltwo{18.10} & \hltwo{15.84} & \underline{30.57} & \hltwo{15.00} & \hltwo{19.35} & \hltwo{24.00} & \hltwo{23.82} \\
\midrule

\rowcolor{modelcolor3} \multicolumn{11}{@{}c}{\textit{Backbone: Qwen3-30B-A3B-Thinking-2507}} \\
{--}       & \xmark & 13.03 & 7.96 & 8.14 & 3.47 & 7.25 & 5.00 & 8.06 & 2.86 & 9.24 \\
{--}       & \cmark & 13.13 & 7.96 & 6.33 & 1.98 & 11.92 & 10.00 & 6.45 & 10.29 & 10.17 \\
TaskCraft  & \cmark & \underline{24.62} & 12.39 & 16.29 & 7.92 & 21.76 & \underline{19.00} & \underline{12.90} & 22.29 & 19.87 \\
MegaScience& \cmark & 23.69 & \underline{14.60} & \underline{20.81} & \underline{9.90} & \underline{26.94} & 15.00 & 8.06 & 18.29 & \underline{20.15} \\
MiroVerse  & \cmark & 23.38 & 12.39 & \underline{20.81} & 9.41 & 24.87 & 7.00 & 11.29 & \underline{22.86} & 19.64 \\
\hlthree{\myname} & \cmark & \hlthree{29.85} & \hlthree{16.81} & \hlthree{21.27} & \hlthree{17.82} & \hlthree{31.61} & \hlthree{22.00} & \hlthree{14.52} & \hlthree{28.00} & \hlthree{25.67} \\
\bottomrule
\end{tabular}%
}
\end{table}
\section{Analysis}

\subsection{BoN Analysis: Validating Difficulty Richness \& Potential for RL Training}

To assess the difficulty distribution of \myname~ and the latent capabilities of the RFT model, we conducted a Best-of-N (BoN) analysis. On a held-out validation set of 300 samples, we generated $N=8$ independent solution trajectories for each task and measured the success rate if at least one of the $N$ attempts was correct (pass@$N$).

\begin{wrapfigure}{r}{0.5\textwidth}
  \vspace{-15pt}
  \centering
  \includegraphics[width=0.48\textwidth]{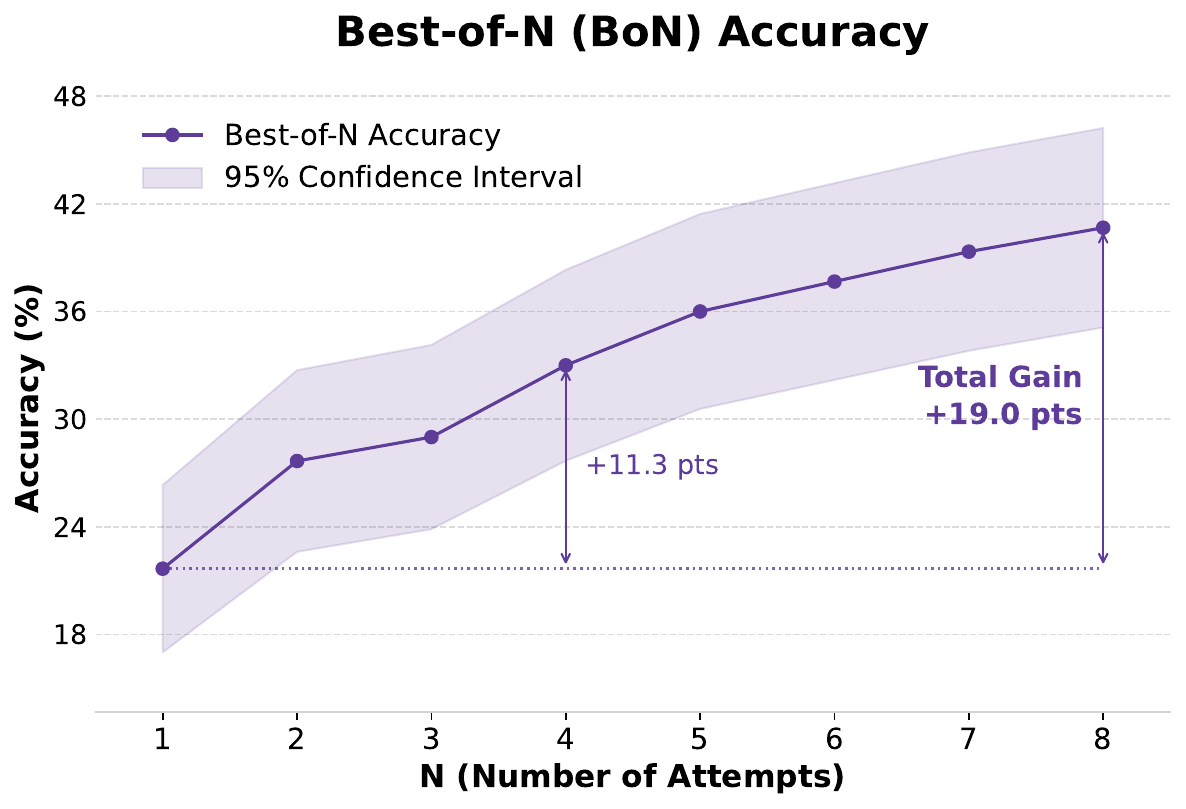} 
  \caption{Best-of-N (BoN) accuracy of our RFT Qwen3-30B-A3B model on a 300-sample validation set from \myname.}
  \label{fig:bo8}
  \vspace{-15pt}
\end{wrapfigure}

As shown in Figure \ref{fig:bo8}, the accuracy dramatically increases from 21.7\% at pass@1 to 40.7\% at pass@8. This 19.0-point improvement provides two key insights. \textbf{First, it validates the designed difficulty of \myname:} the dataset is not a binary mix of trivial and impossible tasks. Instead, it presents a challenging frontier where initial attempts may fail, but success is achievable through exploration. This provides a rich learning signal beyond superficial pattern matching. \textbf{Second, it highlights the significant potential for subsequent reinforcement learning (RL)} While supervised fine-tuning (SFT) trains the model on a single reference solution, the large gap between pass@1 and pass@8 confirms that for problems the model fails to solve on the first attempt, its policy distribution contains diverse and successful alternative trajectories. This is a crucial precondition for effective RL, ensuring that exploration can discover high-reward experiences necessary for effective policy optimization. Therefore, \myname~serves not only as a robust training resources for SFT but also as a strong foundation for RL to further unlock an agent's problem-solving potential.

\subsection{Why \myname~Excels: Deconstructing the Gains in Reasoning and Tool-Use}

\begin{wrapfigure}{r}{0.4\textwidth}
  \vspace{-20pt}
  \centering
  \includegraphics[width=0.4\textwidth]{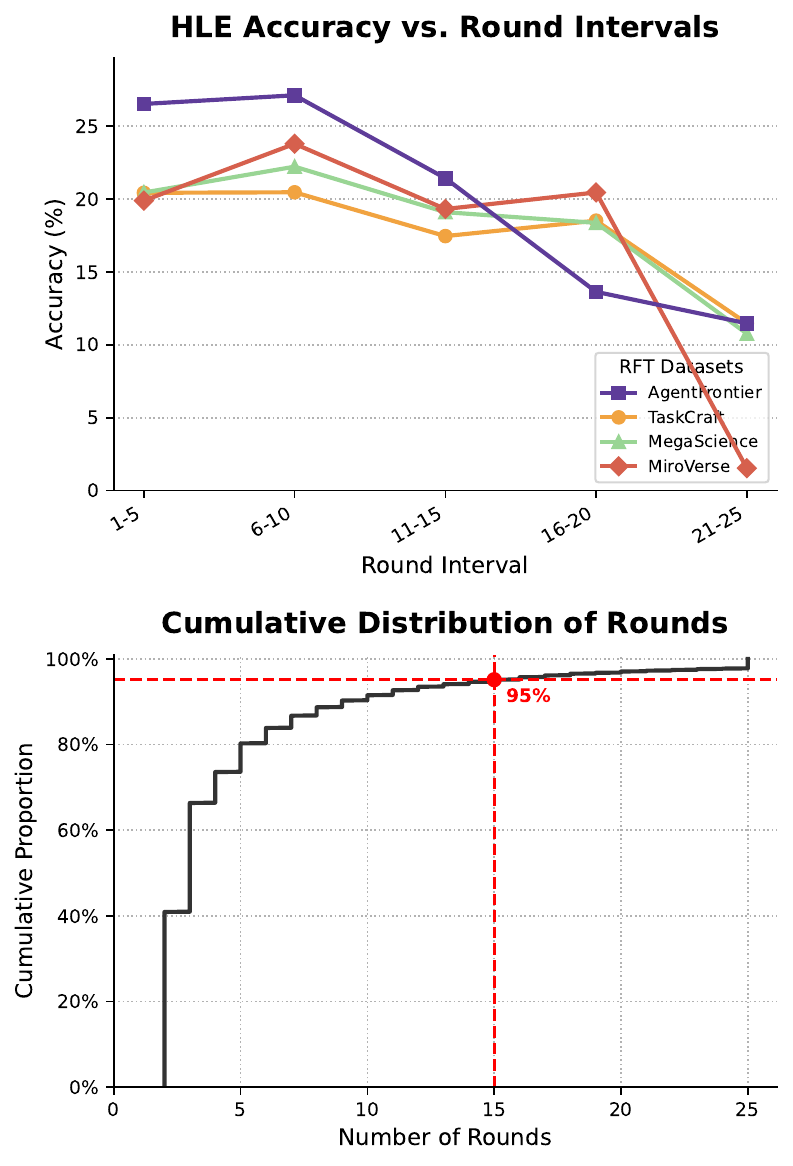} 
  \caption{Accuracy vs. number of rounds on 4 datasets.}
  \label{fig:accuracy_vs_rounds}
  \vspace{-20pt}
\end{wrapfigure}

\textbf{From Shallow Retrieval to Deep Causal Reasoning.} Figure~\ref{fig:accuracy_vs_rounds} reveals the performance dynamics that underscore \myname's superiority. The vast majority (95\%) of problems are solved within a 15-round horizon, a critical window in which our RFT dataset consistently outperforms all fine-tuning dataset baselines. This advantage is a principled consequence of our data generation strategy rooted in the Zone of Proximal Development. By curating tasks that are unsolvable by the base model yet solvable with external scaffolding, we create training instances of optimal difficulty. This forces the model to abandon simplistic, single-source retrieval and instead master knowledge fusion—the non-trivial meta-skill of integrating disparate information streams into a coherent solution. The agent learns not merely what information to retrieve, but how to synthesize it, shifting from shallow pattern-matching to in-depth causal reasoning.

\textbf{From High-Volume Invocation to High-Efficacy Orchestration.} The design philosophy of \myname~prioritizes the cultivation of strategic tool orchestrators over rote tool callers. Unlike datasets that promote skewed tool dependencies (e.g., code-centric MiroVerse or search-centric TaskCraft), \myname~promotes a balanced tool-use distribution (Table~\ref{tab:data_stats}). This forces the agent to develop a sophisticated understanding of inter-tool synergy rather than mastering a single tool in isolation. The results on the HLE benchmark (Table~\ref{tab:tool_analysis}) confirm this empirical payoff. Our agent achieves a macro-average conditional tool accuracy of 26.3\%—a significant leap from the 21\% plateau of competitors—with a comparable number of interactions. This demonstrates that agent capability stems not from the volume of tool calls, but their efficacy. Our method trains the model to transition from high-volume, low-yield tool usage to precise, high-efficacy orchestration, which is a crucial step toward creating more resourceful agents.

\begin{table*}[h]
\centering
\caption{
   Tool usage statistics for the Qwen3-30B-A3B agent on the HLE text-only test set (2154 problems). Each column block shows performance after RFT on a different dataset. We report average usage per round and conditional tool accuracy (Acc, \%), defined as the success rate for tasks that use the tool. The final row details overall metrics. Best results are in \textbf{bold}.
}
\label{tab:tool_analysis}
\begin{tabular}{l *{4}{S[table-format=1.2] S[table-format=2.1]}}
\toprule
& \multicolumn{2}{c}{\textbf{TaskCraft}} & \multicolumn{2}{c}{\textbf{MegaScience}} & \multicolumn{2}{c}{\textbf{MiroVerse}} & \multicolumn{2}{c}{\textbf{\myname}} \\
\cmidrule(lr){2-3} \cmidrule(lr){4-5} \cmidrule(lr){6-7} \cmidrule(lr){8-9}
\textbf{Tool / Metric} & {Usage} & {Acc (\%)} & {Usage} & {Acc (\%)} & {Usage} & {Acc (\%)} & {Usage} & {Acc (\%)} \\
\midrule
Search          & 0.68 & 19.6 & 0.67 & 20.3 &  \textbf{0.73} & 20.4 &  \textbf{0.73} &  \hlthree{24.9} \\
Scholar         & 0.78 & 21.0 &  \textbf{0.98} & 20.3 & 0.87 & 20.6 & 0.89 &  \hlthree{25.4} \\
Browser         & 1.24 & 25.2 & 1.39 & 23.4 &  \textbf{1.47} & 22.7 & 1.32 &  \hlthree{29.8} \\
Code            & 0.52 & 18.1 & 0.65 & 18.6 &  \textbf{0.67} & 18.4 & 0.63 &  \hlthree{24.9} \\
\midrule
\textbf{Overall} (Rounds/Acc.) & 4.21 & 21.0 & 4.70 & 20.6 & \textbf{4.74} & 20.5 & 4.57 & \hlthree{26.3} \\
\bottomrule
\end{tabular}%
\end{table*}

\subsection{Holistic Agentic Training}
\paragraph{Setup} We further investigate the performance gains a holistic training pipeline that incorporates continued pre-training (CPT) and post-training. Due to the large-scale GPU computation in CPT, this study is conducted only on Qwen3-30B-A3B-Thinking-2507 and our \myname~data.
The holistic training pipeline consists of two stages:
\begin{enumerate}
    \item \textbf{Continual Pre-training (CPT)}: One epoch over 50B tokens, comprising 1 million summarized text chunks and 20 million knowledge-intensive QA pairs.;
    \item \textbf{Rejection Sampling Fine-tuning (RFT)}: Three epochs on 12,000 high-quality trajectories.
\end{enumerate}

\paragraph{CPT Objective} The CPT stage minimizes the standard language modeling loss:
\begin{equation}
\mathcal{L}_{\text{CPT}}(\theta) = - \sum_{t=1}^{T} \log p_{\theta}(x_t \mid x_{<t}),
\end{equation}
where $x_t$ denotes the token at position $t$, and $\theta$ are the model parameters.

\begin{table}[h]
\centering
\caption{Comparison of \myname~with state-of-the-art proprietary and open-source LLMs/Agents on four high-level multidisciplinary benchmarks. $^\dagger$ marks the result from the corresponding official reports. The final row highlights the performance gain from our Continual Pre-training (CPT) stage.}
\label{tab:results2}
\resizebox{\textwidth}{!}{%
\begin{tabular}{l|c|cccc}
\toprule
\textbf{LLMs/Agents} & \textbf{Tools} & \textbf{HLE (text-only)}  & \textbf{ZPD Exam-v1} & \textbf{RBench-T} & \textbf{xBench-ScienceQA} \\
\midrule
\rowcolor[RGB]{229,229,252}\multicolumn{6}{c}{\emph{\textbf{Direct Inference (with and without Tools)}}} \\
\midrule
\multirow{2}{*}{GPT-4o} & \xmark & 2.3 &  4.8 & 42.0 & 13.0 \\
                        & \cmark & 4.8 & 51.3 & 48.5 & 15.0 \\
\cmidrule(lr){2-6}
\multirow{2}{*}{Claude 4 Sonnet} & \xmark & 5.4 & 6.0 & 61.8 & 32.0 \\
                                 & \cmark & 14.3 & 86.6 & 71.1 & 47.0 \\
\cmidrule(lr){2-6}
\multirow{2}{*}{Gemini 2.5 Flash} & \xmark & 10.4 & 6.3 & 65.2 & 35.0 \\
                                  & \cmark & 12.6 & 58.1 & 75.8 & 39.0 \\
\cmidrule(lr){2-6}
\multirow{2}{*}{DeepSeek V3.1-671B} & \xmark & 18.5 & 8.2 & 76.3 & 40.0 \\
                                    & \cmark & \textbf{29.8}$^\dagger$ & \underline{93.1} & \textbf{79.4} & \underline{55.0} \\
\cmidrule(lr){2-6}
\multirow{2}{*}{Qwen3-30B-A3B (Thinking-2507)} & \xmark & 9.2 & 4.9 & 51.2 & 32.0 \\
                                             & \cmark & 10.2 & 47.2 & 55.1 & 40.0 \\
\midrule
\rowcolor[RGB]{229,229,252}\multicolumn{6}{c}{\emph{\textbf{Proprietary Research Agents}}} \\
\midrule
OpenAI DeepResearch & \cmark & 26.6$^\dagger$ & -- & -- & -- \\
Gemini DeepResearch & \cmark & 26.9$^\dagger$ & -- & -- & -- \\
Kimi-Researcher & \cmark & 26.9$^\dagger$ & -- & -- & -- \\
\midrule
\rowcolor[RGB]{229,229,252}\multicolumn{6}{c}{\emph{\textbf{Open-source Agents}}} \\
\midrule
WebDancer-QwQ-32B & \cmark & 6.4 & 51.8 & 67.6 & 38.0 \\
WebSailor-72B & \cmark & 9.2 & 62.1 & 44.9 & 27.0 \\
WebShaper-72B & \cmark & 8.0 & 54.4 & 66.8 & 29.0 \\
\midrule
\rowcolor[RGB]{229,229,252}\multicolumn{6}{c}{\emph{\textbf{Ours}}} \\
\midrule
\textbf{\myname-30B-A3B (RFT only)} & \cmark & 25.7 & 91.4 & 74.4 & 54.0 \\
\textbf{\myname-30B-A3B (CPT+RFT)} & \cmark & \underline{28.6} & \textbf{93.4} & \underline{77.1} & \textbf{61.0} \\
\cmidrule(lr){1-6} 
\rowcolor{gray!15} 
\textbf{$\Delta$ (CPT gain)} & & \textbf{+2.9} & \textbf{+2.0} & \textbf{+2.7} & \textbf{+7.0} \\
\bottomrule
\end{tabular}%
}
\end{table}

\paragraph{Evaluation} To comprehensively assess our model, \myname~(CPT+RFT), we conduct extensive evaluations against a diverse range of competitors. These include leading closed-source~\citep{gpt4o, claude, gemini2.5} and open-source~\citep{deepseekv3, yang2025qwen3} language models, evaluated with and without access to external tools. Additionally, we compare \myname~with both proprietary deep-research agents~\citep{dr, google_dr, kimi-researcher} and prominent open-source agents~\citep{wu2025webdancer, li2025websailor, tao2025webshaper}. 

\paragraph{Main Results} Table~\ref{tab:results2}, our holistically trained agent not only sets a new state-of-the-art among open-source models but also competes effectively with significantly larger, proprietary agents. The final row isolates the contribution of CPT, which consistently boosts performance across all benchmarks (+2.9 on HLE, +7.0 on xBench-ScienceQA). Notably, CPT yields a +2.0 point gain on ZPD Exam, where the RFT-only model's performance was already near-saturation. This provides strong evidence that strengthening a model's foundational knowledge via CPT directly enhances its capacity for complex agentic tasks.

\subsection{Case Study}
\definecolor{mygreen}{RGB}{0, 128, 0} 
A qualitative analysis on an HLE case~\citep{hle} (Appendix \ref{app:case_study}) further illustrates our agent's reasoning process. In a complex clinical scenario, OpenAI DeepResearch~\citep{dr} agent exhibited \textbf{diagnostic fixation}, misdiagnosing \textcolor{red}{\textit{\textbf{Charcot Arthropathy}}} by focusing on common negative findings like sterile synovial fluid. In contrast, our \myname~agent correctly identified the key anomaly: the patient's paradoxical worsening on prednisone. It hypothesized that this was due to a latent infection unmasked by immunosuppression, rather than an inflammatory rebound. This triggered a targeted inquiry, using a literature search to confirm that \textcolor{mygreen}{\textit{\textbf{Chronic Osteomyelitis}}} can present with sterile aspirates and is exacerbated by steroids. This progression from identifying an anomaly to forming a hypothesis and validating it with targeted research demonstrates \myname's advanced research capabilities.

\section{Related Work}

\paragraph{Data Synthesis for LLM Agents} 
Synthesizing high-quality data is critical for advancing LLM agents that require complex reasoning and tool use~\citep{zeng2025glm, liu2025synlogic, zhou2024jiuzhang3}. Initial efforts replaced costly manual curation with programmatic generation, creating agentic tasks with verifiable solution trajectories~\citep{shi2025taskcraft, hongjinlearn, huang2024datagen}. Subsequent research aimed to enhance data quality by grounding synthesis in external knowledge sources like scientific documents~\citep{fan2025megascience, fengbeyond}. While these approaches increase factual richness, they often produce tasks solvable via localized information retrieval, rather than promoting the deep knowledge integration essential for complex research~\citep{dr}. A central challenge remains the precise calibration of task difficulty. Without a principled control mechanism, synthetic data risks being too simple for effective learning or too complex to yield a usable training signal~\citep{limontessori}. These strategies rely on heuristics like incremental constraint addition~\citep{patel2025get} or probes to distinguish reasoning from recitation~\citep{yan2025recitation}, yet lack a principled framework to calibrate difficulty for scaffolding complex reasoning.

\paragraph{Multi-disciplinary Benchmark} The evaluation of advanced reasoning in large language models (LLMs) was pioneered by MMLU~\citep{mmlu}, which set the standard for assessing multi-disciplinary knowledge. This led to a wave of subsequent benchmarks~\citep{gpqa, mmlupro, supergpqa, guorbench, xbench} targeting undergraduate or graduate level knowledge. However, the rapid progress of frontier models~\citep{o3, gemini2.5, claude} is causing performance saturation on these static benchmarks, reducing their effectiveness in differentiating top-tier models. While newer benchmarks like Humanity's Last Exam~\citep{hle} increase difficulty through expert curation, they remain fixed assessments. In contrast, our work introduces the ZPD Exam, a self-evolving evaluation framework that adapts in lockstep with model capabilities, providing a consistently challenging frontier for LLM agent evaluation.

\paragraph{Deep-Research Agents} Deep-research agent, a system built upon large reasoning models (LRMs), is designed to automate multi-step search and reasoning. It empowers users to complete complex, cross-domain information synthesis and in-depth research tasks in minutes, a process that would otherwise require hours of human effort. Proprietary agents~\citep{dr, google_dr, claude_deep_research, grok3, Perplexity, kimi-researcher} have demonstrated impressive capabilities in complex, multi-step research tasks. The open-source community has fostered a rich ecosystem of transparent and reproducible agents~\citep{jin2025search,li2025search,Li2025webthinker,tao2025webshaper,li2025websailor, qiao2025webresearcher}. These efforts typically leverage explicit planning, tool-use, and web navigation to emulate human research processes, advancing the field through shared methodologies.

\section{Conclusion}

In this work, we presented a novel data synthesis paradigm based on the Zone of Proximal Development (ZPD) theory. Our framework co-generates a targeted training resources and a self-evolving ZPD Exam to progressively enhance and evaluate agentic reasoning. The resulting model, \myname-30B-A3B, validates our approach by achieving state-of-the-art results on challenging expert-level multi-disciplinary benchmarks, surpassing even significantly larger proprietary agents. This work demonstrates that a principled, pedagogical approach to data synthesis is a highly effective, if not essential, strategy for cultivating advanced reasoning abilities in a data-efficient manner.

\section*{Limitations and Future Work}
While our ZPD-guided framework demonstrates significant promise, we identify three primary limitations that chart clear paths for future research: 

\begin{enumerate}
    \item \textbf{Graduated Scaffolding:} Our current ZPD operationalization relies on binary, "all-or-nothing" scaffolding, where the More Knowledgeable Other (MKO) provides a complete solution trajectory. This simplifies the nuanced support common in human pedagogy. A key future direction is to develop graduated scaffolding, offering tiered assistance from high-level strategic hints to specific sub-goals. Such a system would not only teach the agent what to do with help but also foster the crucial meta-cognitive skill of learning how to seek it, leading to more autonomous and sample-efficient learning.
    \item \textbf{From Imitation to Exploration:} Our reliance on imitation learning (IL), specifically Rejection-Sampling Fine-Tuning, constrains the agent to mode-seeking behavior. The significant gap between our pass@1 and pass@N results strongly indicates a diverse distribution of valid solutions that IL under-explores. This presents a prime opportunity for Reinforcement Learning (RL). We propose using our fine-tuned model as a high-quality policy prior to initialize an RL agent, and repurposing the ZPD-guided data as a principled reward signal. This shift from imitation to exploration would empower the agent to discover novel and superior policies, breaking beyond the performance ceiling of the demonstration data.
    \item \textbf{Dynamic Tool Creation:} The agent's problem-solving capacity is currently bounded by its predefined, static toolset. While proficient as a tool user, it cannot function as a tool creator. A pivotal advancement is to empower the agent with tool creation abilities, pursuing two complementary paths: (1) Hierarchical Tool Composition, learning to combine existing tools into reusable "meta-tools" for recurring sub-tasks; and (2) Program Synthesis, programmatically generating new functions to address novel problem requirements. This evolution from tool user to creator is a critical step towards more general and resourceful agents capable of dynamically extending their capabilities for a broader problem space.
\end{enumerate}

\section*{Acknowledgment}
We sincerely thank Kuan Li for providing the LaTeX template used in the preparation of this paper.

\clearpage
\appendix
\section{Data Engine Details}
\label{apx:data_engine}

\begin{algorithm}[ht!]
\caption{\myname~Data Engine Pipeline}
\label{alg:data_engine_revised}
\begin{algorithmic}[1]
\Statex \textbf{Input:}
\Statex \quad $\mathcal{C}_{\text{raw}}$: Raw document corpus; $\Phi_{\text{chunk}}$: Chunking model; $\mathcal{M}_{\text{gen}}, \mathcal{A}_{\text{refine}}, \mathcal{A}_{\text{LKP}}, \mathcal{A}_{\text{MKO}}$: Models and agents; $\text{Sim}, \text{IsCorrect}, \text{IsSolvableBy}$: Similarity and evaluation functions; $\tau_{\text{theme}}, K, N, \epsilon, k_{\text{nn}}$: Hyperparameters (thematic threshold, escalation steps, BoN size, redundancy threshold, number of neighbors)

\Statex \textbf{Output:}
\Statex \quad $\mathcal{D}_{\text{ZPD}}$: Calibrated training dataset for post-training; $\mathcal{D}_{\text{pretrain}}$: Dataset for continued pre-training; $\mathcal{D}_{\text{human}}$: Dataset for human review

\Statex
\Procedure{GenerateZPDData}{$\mathcal{C}_{\text{raw}}, \dots$}
    \State $\mathcal{D}_{\text{ZPD}}, \mathcal{D}_{\text{pretrain}}, \mathcal{D}_{\text{human}} \gets \emptyset, \emptyset, \emptyset$
    \Statex
    \Comment{\textbf{Stage I: Seed Question Generation}}
    \State $\mathcal{C}_{\text{chunk}} \gets \bigcup_{d \in \mathcal{C}_{\text{raw}}} \Phi_{\text{chunk}}(d)$ \Comment{Preprocess corpus into semantic chunks}
    \State $\mathcal{V}_{\text{index}} \gets \text{BuildVectorIndex}(\mathcal{C}_{\text{chunk}})$ \Comment{Build index for efficient search}
    \State $\mathcal{D}_{\text{seed}} \gets \emptyset$
    \For{each chunk $c_i \in \mathcal{C}_{\text{chunk}}$}
        \State $\mathcal{N}_i \gets \text{FindNearestNeighbors}(c_i, \mathcal{V}_{\text{index}}, k_{\text{nn}})$ \Comment{Find k-NN for efficient combination}
        \For{each pair $(c_j, c_k)$ from $\mathcal{N}_i$}
            \If{$\text{Sim}(c_i, c_j) > \tau_{\text{theme}} \land \text{Sim}(c_i, c_k) > \tau_{\text{theme}} \land \text{Sim}(c_j, c_k) > \tau_{\text{theme}}$}
                \State $(q_0, a_0) \gets \mathcal{M}_{\text{gen}}(\{c_i, c_j, c_k\})$ \Comment{Generate QA from thematic unit}
                \State $\mathcal{D}_{\text{seed}} \gets \mathcal{D}_{\text{seed}} \cup \{(q_0, a_0)\}$
            \EndIf
        \EndFor
    \EndFor
    \Statex
    \Comment{\textbf{Stages II \& III: Iterative Escalation and ZPD Calibration}}
    \State $\mathcal{V}_{\text{ZPD}} \gets \text{BuildVectorIndex}(\emptyset)$ \Comment{Initialize index for ZPD-set diversity check}
    \For{each $(q_0, a_0)$ in $\mathcal{D}_{\text{seed}}$}
        \State $(q, a) \gets (q_0, a_0)$
        \Statex \quad \Comment{\textbf{Stage II: Agentic Refinement}}
        \For{$k=1$ to $K$} \Comment{Iteratively escalate complexity}
             \State $(q, a) \gets \Psi_{\text{escalate}}(q, a, \mathcal{A}_{\text{refine}})$ \Comment{e.g., Expand, Abstract, Ground, etc.}
        \EndFor
        \Statex \quad \Comment{\textbf{Stage III: ZPD-based Filtering}}
        \If{$\text{IsSolvableBy}(\mathcal{A}_{\text{LKP}}, q, a)$} \Comment{Check if too easy for Less Knowledgeable Peer}
            \State $\mathcal{D}_{\text{pretrain}} \gets \mathcal{D}_{\text{pretrain}} \cup \{(q, a)\}$
        \Else \Comment{Challenging for LKP, now verify with MKO}
            \State $S_{\text{solutions}} \gets \{ \mathcal{A}_{\text{MKO}}(q) \text{ for } i=1 \dots N \}$ \Comment{Best-of-N by More Knowledgeable Other}
            \If{$\exists s \in S_{\text{solutions}} \text{ s.t. } \text{IsCorrect}(s, a)$} \Comment{Verified as solvable, thus within ZPD}
                \State $q_{\text{nearest}} \gets \text{FindNearestNeighbor}(q, \mathcal{V}_{\text{ZPD}})$
                \If{$q_{\text{nearest}} = \emptyset$ or $\text{Sim}(q, q_{\text{nearest}}) < \epsilon$} \Comment{Filter for diversity}
                    \State $\mathcal{D}_{\text{ZPD}} \gets \mathcal{D}_{\text{ZPD}} \cup \{(q, a)\}$
                    \State $\text{UpdateVectorIndex}(\mathcal{V}_{\text{ZPD}}, q)$
                \EndIf
            \Else \Comment{Unsolvable by MKO, potentially flawed or too hard}
                \State $\mathcal{D}_{\text{human}} \gets \mathcal{D}_{\text{human}} \cup \{(q, a)\}$
            \EndIf
        \EndIf
    \EndFor
    \State \textbf{return} $\mathcal{D}_{\text{ZPD}}, \mathcal{D}_{\text{pretrain}}, \mathcal{D}_{\text{human}}$
\EndProcedure
\end{algorithmic}
\end{algorithm}

This section provides a detailed breakdown of the hyperparameters, procedural logic, and computational costs associated with the \myname~Data Engine, as outlined in Algorithm~\ref{alg:data_engine_revised}. These details are provided to ensure the transparency and reproducibility of our data synthesis framework.

\subsection{Hyperparameter Configuration}

The data generation pipeline is governed by several key hyperparameters that control the granularity of data sourcing, the complexity of generated questions, and the strictness of the filtering process. Our configuration is as follows:

\begin{itemize}
    \item \textbf{Thematic Coherence Threshold ($\tau_{\text{theme}}$):} Set to \textbf{0.8}. This value determines the minimum semantic similarity required between text chunks to form a "composite unit" for seed question generation. A higher value ensures that initial questions are synthesized from thematically tighter content, promoting knowledge fusion.
    \item \textbf{Nearest Neighbors for Seeding ($k_{\text{nn}}$):} Set to \textbf{10}. During seed generation, for each text chunk, we retrieve its $k_{\text{nn}}$ nearest neighbors to search for coherent triplets. This balances computational efficiency with a sufficiently large search space for discovering novel combinations.
    \item \textbf{Maximum Refinement Iterations ($K_{\text{max}}$):} Set to \textbf{30}. This parameter defines the maximum number of complexity escalation steps for any given QA pair in Stage II. This upper bound prevents infinite loops and manages computational resources.
    \item \textbf{Best-of-N (BoN) Verification Size ($N$):} Set to \textbf{3}. In the ZPD-filtering stage, the More Knowledgeable Other ($\mathcal{A}_{\text{MKO}}$) makes $N$ independent attempts to solve a problem. This helps to reduce the variance in the agent's performance and provides a more reliable signal of whether a task is solvable.
    \item \textbf{Diversity Filter Threshold ($\epsilon$):} Set to \textbf{0.7}. To ensure dataset diversity, a new QA pair is discarded if its question's semantic similarity to any existing question in $\mathcal{D}_{\text{ZPD}}$ exceeds this threshold. The similarity is measured by a state-of-the-art reranker model.
\end{itemize}

\subsection{Agentic Refinement and Stopping Criterion}

The core of our data engine is the iterative refinement loop (Stage II), driven by the agent $\mathcal{A}_{\text{refine}}$. The goal of the escalation operator, $\Psi_{\text{escalate}}$, is to progressively increase the cognitive load required to answer a question. This is achieved by prompting the agent to perform a series of enrichment actions, including but not limited to: expanding the question with new, relevant concepts discovered through tool use; abstracting a general principle from specific examples; grounding the problem in a more complex, realistic context; or transforming a qualitative problem into a quantitative one requiring computation.

The iterative escalation is guided by a principled stopping criterion tied to the ZPD framework: for a given QA pair, the refinement loop terminates when the generated question $q_k$ becomes unsolvable by the \textbf{Less Knowledgeable Peer} ($\mathcal{A}_{\text{LKP}}$), a baseline model formally defined in Stage III, or when a predefined maximum of $K_{\text{max}}=30$ iterations is reached. This targeted termination ensures that the engine's computational resources are focused on producing problems that precisely challenge the base model's capabilities.

\subsection{Computational Cost Analysis}

We provide a detailed analysis of the computational cost required to generate a single high-quality data point for the $\mathcal{D}_{\text{ZPD}}$ dataset. The cost is broken down into the two primary stages of our pipeline: agentic refinement and MKO verification. All token counts are based on the respective model's tokenizer, and costs are estimated using official API pricing as of the experiment date\footnote{Pricing references: DeepSeek Model API (\url{https://api-docs.deepseek.com/}), SerpApi for Google Search (\url{https://serpapi.com/enterprise}), and Jina Reader API (\url{https://jina.ai/reader/})}.

\subsubsection{Cost of Agentic Refinement (Stage II)}
In this stage, the refinement agent, $\mathcal{A}_{\text{refine}}$, iteratively enhances a QA pair until it reaches the capability frontier of the Less Knowledge Peer (LKP). The cost per data point is variable, depending on the number of iterations ($K$) needed.

On average, processing a single candidate data point involves the following:
\begin{itemize}
    \item \textbf{Refinement Iterations ($K$):} A data point undergoes an average of \textbf{7.81} iterations.
    \item \textbf{Token Throughput per API Call:}
    \begin{itemize}
        \item Input: \textbf{18,613.82} tokens.
        \item Output: \textbf{11,643.22} tokens.
    \end{itemize}
    \item \textbf{Tool Calls per Data Point:}
    \begin{itemize}
        \item Search: \textbf{0.70} calls.
        \item Scholar: \textbf{0.61} calls.
        \item Browser: \textbf{1.21} calls (avg. 10,000 tokens/call).
        \item Code Interpreter: \textbf{0.94} calls (executed locally, no API cost).
    \end{itemize}
\end{itemize}

\paragraph{Cost Breakdown.} The average refinement cost per candidate is approximately \textbf{\$0.24}, calculated as follows:
\begin{itemize}
    \item \textbf{LLM Cost:} $7.81 \times (18,614 \times \$0.56/\text{M} + 11,643 \times \$1.68/\text{M}) \approx \$0.234$.
    \item \textbf{Search Cost:} $(0.70 + 0.61) \times \$0.00275/\text{call} \approx \$0.0036$.
    \item \textbf{Browser Cost:} $1.21 \times 10,000 \times \$0.00005/\text{token} \approx \$0.0006$.
\end{itemize}

\subsubsection{Cost of MKO Verification (Stage III)}
Candidates that pass the refinement stage are then verified by the More Knowledgeable Other agent, $\mathcal{A}_{\text{MKO}}$. This Best-of-N ($N=3$) verification confirms that the problem is solvable by an expert-level agent, thus ensuring its placement within the Zone of Proximal Development (ZPD).

For the $N=3$ verification attempts on a single candidate, the average resource consumption is:
\begin{itemize}
    \item \textbf{Total API Calls:} \textbf{3.32} calls.
    \item \textbf{Token Throughput per API Call:}
    \begin{itemize}
        \item Input: \textbf{20,181.57} tokens.
        \item Output: \textbf{24,169.88} tokens.
    \end{itemize}
    \item \textbf{Total Tool Calls:}
    \begin{itemize}
        \item Search: \textbf{0.50} calls.
        \item Scholar: \textbf{0.92} calls.
        \item Browser: \textbf{1.30} calls (avg. 10,000 tokens/call).
        \item Code Interpreter: \textbf{0.53} calls (executed locally, no API cost).
    \end{itemize}
\end{itemize}

\paragraph{Cost Breakdown.} The verification cost for a single candidate is approximately \textbf{\$0.18}:
\begin{itemize}
    \item \textbf{LLM Cost:} $3.32 \times (20,182 \times \$0.56/\text{M} + 24,170 \times \$1.68/\text{M}) \approx \$0.172$.
    \item \textbf{Search Cost:} $(0.50 + 0.92) \times \$0.00275/\text{call} \approx \$0.0039$.
    \item \textbf{Browser Cost:} $1.30 \times 10,000 \times \$0.00005/\text{token} \approx \$0.00065$.
\end{itemize}
However, only a fraction of candidates pass this stage. With an observed success rate of \textbf{33\%}, the amortized cost to obtain one successfully verified data point is $\$0.18 / 0.33 \approx \textbf{\$0.54}$.

In summary, the total end-to-end amortized cost to generate one high-quality, verified PhD-level QA pair with its solution trajectory for $\mathcal{D}_{\text{ZPD}}$ is approximately \textbf{\$0.78} (\$0.24 for refinement + \$0.54 for amortized verification). While this represents a non-trivial investment per sample, it aligns with our "quality-over-quantity" approach. This automated pipeline produces a valuable training asset at a fraction of the cost and time that manual curation by human experts would demand.

\section{Experimental Details}
\label{apx:exp}
\subsection{Tools Implementation}
\label{apx:tool}
Our agent is equipped with a suite of tools to support its research process, from broad exploration to empirical validation. Each tool is designed for batch processing to enhance efficiency and produces structured outputs for seamless integration into the agent's iterative reasoning loop.
\begin{itemize}
    \item \textbf{Search:} Performs parallel web searches using the Google Search API. It returns a list of structured results, each containing a title, snippet, and URL, allowing the agent to efficiently assess the relevance of multiple sources.
    \item \textbf{Scholar:} Tackles multi-disciplinary challenges by querying the Google Scholar API to navigate scientific literature. It returns structured metadata, including authors, publication venue, and citation counts, enabling the agent to identify authoritative works and their scholarly context.
    \item \textbf{Browser:} Extracts targeted information from a given URL. The agent provides a specific goal (e.g., "extract the dataset and evaluation metrics"). The tool first fetches the page content using Jina Reader~\citep{jina} and then employs Qwen3~\citep{yang2025qwen3} to synthesize a precise answer based on the goal. This allows for focused knowledge extraction from web pages. 
    \item \textbf{Code:} Provides a sandboxed Python environment for computational analysis and verification. It is equipped with standard scientific libraries (e.g., NumPy, SciPy) and allows the agent to execute code for tasks like data analysis or simulations. All outputs (stdout, stderr, and figures) are captured as text, providing empirical evidence for the agent's reasoning process.
\end{itemize}

\subsection{Training Details}
We implement supervised fine-tuning (SFT) using the Megatron-LM framework~\citep{shoeybi2019megatron}. The hyperparameters for fine-tuning our MoE and Dense models are detailed in Table~\ref{tab:sft_moe_params} and Table~\ref{tab:sft_dense_params}, respectively.

\begin{table}[h!]
\centering
\begin{minipage}{0.45\textwidth}
    \centering
    \caption{SFT Hyperparameters for the MoE Model.}
    \label{tab:sft_moe_params}
    \resizebox{\textwidth}{!}{%
    \begin{tabular}{@{}lr@{}}
        \toprule
        \textbf{Parameter} & \textbf{Value} \\
        \midrule
        Training Epochs        & 3 \\
        Max Sequence Length    & 40,960 \\
        Batch Size      & 256 \\
        Learning Rate    & $7.0 \times 10^{-6}$ \\
        Learning Rate (Min)    & $7.0 \times 10^{-7}$ \\
        LR Scheduler           & Linear Decay \\
        Tensor Parallel (MP)   & 4 \\
        Expert Parallel (EP)   & 2 \\
        Pipeline Parallel (PP) & 1 \\
        \bottomrule
    \end{tabular}%
    }
\end{minipage}\hfill
\begin{minipage}{0.45\textwidth}
    \centering
    \caption{SFT Hyperparameters for the Dense Model.}
    \label{tab:sft_dense_params}
    \resizebox{\textwidth}{!}{%
    \begin{tabular}{@{}lr@{}}
        \toprule
        \textbf{Parameter} & \textbf{Value} \\
        \midrule
        Training Epochs        & 3 \\
        Max Sequence Length    & 40,960 \\
        Batch Size  & 64 \\
        Learning Rate          & $4.0 \times 10^{-5}$ \\
        LR Scheduler           & Cosine Decay \\
        Warmup Ratio           & 0.1 \\
        \bottomrule
    \end{tabular}%
    }
\end{minipage}
\end{table}

\subsection{More Results on on Fine-tuning Datasets}
Table \ref{tab:tool_analysis_modified} presents a detailed analysis of tool usage and conditional accuracy for Qwen3-30B-A3B model after undergoing rejection-sampling fine-tuning (RFT) on four distinct datasets. The results clearly demonstrate the effectiveness of our synthesized dataset, \myname. The agent fine-tuned on \myname{} achieves the highest overall conditional accuracy on both the ZPD-Exam (87.6\%) and RBench-T (63.7\%) benchmarks. Furthermore, it consistently secures top-tier accuracy for critical tools across various benchmarks, such as for the Scholar (91.7\%) and Browser (91.8\%) tools on ZPD-Exam and the Code tool on both ZPD-Exam (83.3\%) and RBench-T (78.6\%). This superior performance underscores the quality of \myname{} in enhancing an agent's capability to correctly and robustly utilize tools across a diverse range of complex tasks.

\begin{table*}[h]
\scriptsize
\centering
\caption{
   Tool usage statistics for the Qwen3-30B-A3B agent on the ZPD Exam, RBench-T and xBench-ScienceQA. Each column block shows performance after RFT on a different dataset. We report average usage per round and conditional tool accuracy (Acc, \%), defined as the success rate for tasks that use the tool. The final row details overall metrics. Best results are in \textbf{bold}.
}
\label{tab:tool_analysis_modified} 
\begin{tabular}{ll *{4}{S[table-format=1.2] S[table-format=2.1]}}
\toprule
& \textbf{Fine-tuning Dataset} & \multicolumn{2}{c}{\textbf{TaskCraft}} & \multicolumn{2}{c}{\textbf{MegaScience}} & \multicolumn{2}{c}{\textbf{MiroVerse}} & \multicolumn{2}{c}{\textbf{\myname}} \\
\cmidrule(lr){3-4} \cmidrule(lr){5-6} \cmidrule(lr){7-8} \cmidrule(lr){9-10}
\textbf{Benchmark} & \textbf{Tool / Metric} & {Usage} & {Acc (\%)} & {Usage} & {Acc (\%)} & {Usage} & {Acc (\%)} & {Usage} & {Acc (\%)} \\
\midrule

\multirow{5}{*}{HLE} 
& Search          & 0.68 & 19.6 & 0.67 & 20.3 & \textbf{0.73} & 20.4 & \textbf{0.73} & \hlthree{24.9} \\
& Scholar         & 0.78 & 21.0 & \textbf{0.98} & 20.3 & 0.87 & 20.6 & 0.89 & \hlthree{25.4} \\
& Browser         & 1.24 & 25.2 & 1.39 & 23.4 & \textbf{1.47} & 22.7 & 1.32 & \hlthree{29.8} \\
& Code            & 0.52 & 18.1 & 0.65 & 18.6 & \textbf{0.67} & 18.4 & 0.63 & \hlthree{24.9} \\
\cmidrule(lr){2-10}
& \textbf{Overall} (Rounds/Acc.) & 4.21 & 21.0 & 4.70 & 20.6 & \textbf{4.74} & 20.5 & 4.57 & \hlthree{26.3} \\
\midrule

\multirow{5}{*}{ZPD-Exam} 
& Search          & 0.15 & \hlthree{90.8} & 0.10 & 85.4 & \textbf{0.18} & 74.8 & 0.13 & 83.6 \\
& Scholar         & 1.20 & 90.1 & \textbf{1.28} & 90.2 & 1.22 & 87.3 & 1.23 & \hlthree{91.7} \\
& Browser         & 1.39 & 90.6 & 1.35 & 91.0 & \textbf{1.46} & 86.9 & 1.45 & \hlthree{91.8} \\
& Code            & 0.03 & 78.1 & 0.03 & 68.6 & 0.02 & 66.7 & \textbf{0.04} & \hlthree{83.3} \\
\cmidrule(lr){2-10}
& \textbf{Overall} (Rounds/Acc.) & 3.77 & 87.4 & 3.76 & 83.8 & \textbf{3.88} & 78.9 & 3.84 & \hlthree{87.6} \\
\midrule

\multirow{5}{*}{RBench-T} 
& Search          & 0.23 & 55.0 & 0.24 & 53.6 & 0.26 & 50.0 & \textbf{0.28} & \hlthree{58.1} \\
& Scholar         & 0.14 & \hlthree{63.1} & 0.15 & 59.6 & \textbf{0.16} & 54.8 & \textbf{0.16} & 59.7 \\
& Browser         & 0.20 & 54.4 & 0.22 & 53.8 & \textbf{0.28} & 46.9 & 0.27 & \hlthree{58.2} \\
& Code            & 0.74 & 77.5 & 0.80 & \hlthree{78.6} & 0.83 & 77.2 & \textbf{0.88} & \hlthree{78.6} \\
\cmidrule(lr){2-10}
& \textbf{Overall} (Rounds/Acc.) & 2.31 & 62.5 & 2.41 & 61.4 & 2.53 & 57.2 & \textbf{2.59} & \hlthree{63.7} \\
\midrule

\multirow{5}{*}{xBench-SciQA} 
& Search          & \textbf{0.44} & 28.6 & 0.39 & 50.0 & 0.36 & 46.4 & 0.43 & \hlthree{57.1} \\
& Scholar         & 0.29 & 54.2 & \textbf{0.39} & 44.8 & 0.36 & \hlthree{66.7} & 0.28 & 48.1 \\
& Browser         & 0.46 & 31.6 & \textbf{0.61} & 38.5 & 0.48 & \hlthree{52.4} & 0.36 & 42.1 \\
& Code            & \textbf{0.62} & 47.2 & 0.54 & 46.8 & 0.60 & 42.6 & 0.58 & \hlthree{55.6} \\
\cmidrule(lr){2-10}
& \textbf{Overall} (Rounds/Acc.) & 2.81 & 40.4 & \textbf{2.93} & 45.0 & 2.81 & \hlthree{52.0} & 2.66 & 50.7 \\
\bottomrule
\end{tabular}
\vspace{-10pt}
\end{table*}

\newpage
\section{Case Study}
\label{app:case_study}

\definecolor{myblue}{RGB}{4, 104, 173}      
\definecolor{mybluebg}{RGB}{232, 242, 250}   
\definecolor{myred}{RGB}{192, 0, 0}         
\definecolor{myredbg}{RGB}{253, 233, 233}     
\definecolor{mygreen}{RGB}{0, 128, 0}       
\definecolor{mygreenbg}{RGB}{235, 248, 235}   
\definecolor{mygray}{RGB}{100, 100, 100}     

\newtcolorbox{problembox}[1]{
    enhanced, breakable,
    colback=mybluebg,
    colframe=myblue,
    coltitle=white,
    fonttitle=\bfseries,
    attach boxed title to top center={yshift=-2mm},
    boxed title style={
        colback=myblue,
        sharp corners,
    },
    title=#1,
}

\newtcolorbox{competitorbox}[1]{
    enhanced, breakable,
    colback=myredbg,
    colframe=myred,
    coltitle=black,
    fonttitle=\bfseries,
    title=#1,
}

\newtcolorbox{ourmodelbox}[1]{
    enhanced, breakable,
    colback=mygreenbg,
    colframe=mygreen,
    coltitle=black,
    fonttitle=\bfseries,
    title=#1,
}

\newtcolorbox{roundbox}[1]{
    colback=gray!15,
    colframe=gray!75,
    fonttitle=\bfseries,
    coltitle=black,
    center title,
    sharp corners,
    boxrule=0.5pt,
    top=2pt, bottom=2pt,
    title=#1,
}

\newtcolorbox{answerbox}{
    colback=green!5!white,
    colframe=green!75!black,
    fonttitle=\bfseries,
    title=Final Answer and Clinical Reasoning,
    sharp corners,
    breakable,
}

\lstdefinestyle{jsonstyle}{
    basicstyle=\ttfamily\small,
    backgroundcolor=\color{black!5},
    showstringspaces=false,
    breaklines=true,
    breakatwhitespace=true,
    frame=tb,
    framerule=0.5pt,
    rulecolor=\color{gray!50},
    keywordstyle=\color{blue},
    stringstyle=\color{red!60!black},
    commentstyle=\color{green!50!black},
    numbers=none,
    literate=
      {<}{{\textless}}1
      {>}{{\textgreater}}1
      {"}{{\textquotedbl}}1,
}

\begin{problembox}{Clinical Scenario from Humanity's Last Exam}
    \textbf{Question:} A 68 year old patient presents to your clinic with ankle pain and swelling. The swelling began after a long walk with his wife the previous day. Physical exam shows erythema, slight edema, and pain on flexion and extension as well as mild bony tenderness. X-rays are negative for any acute abnormality. The patient is treated with indomethacin. He returns to the clinic 10 days later with the same pain, swelling, and redness. Laboratory workup reveals a slightly elevated uric acid and C-reactive protein. X-rays once again rule out any acute findings. The patient is treated with a prednisone taper, but his symptoms begin to worsen. Joint aspiration and synovial fluid analysis reveals no crystals and gram stain reveals no organisms or white blood cells. What is this patient's diagnosis?
    
    \vspace{1em}
    \hrule
    \vspace{1em}
    
    \noindent\textbf{Ground Truth Diagnosis:} \textcolor{mygreen}{\textbf{Chronic osteomyelitis}}
\end{problembox}

\vspace{1.5em} 

\begin{competitorbox}{\textcolor{white}{\faTimesCircle\ \ OpenAI DeepResearch Output (Incorrect)}}
Based on the clinical presentation and diagnostic findings, this patient's most likely diagnosis is \textcolor{red}{\textbf{Charcot Arthropathy}}. Here’s the reasoning:
\begin{enumerate}
    \item Chronic and Recurrent Swelling with Erythema: Charcot arthropathy often presents with recurrent swelling, warmth, and erythema, which can be mistaken for infections or inflammatory arthritis.
    \item X-rays Negative for Acute Findings: Early Charcot arthropathy may not show significant abnormalities on X-rays, though later stages may demonstrate joint destruction or deformity.
    \item Joint Aspiration Findings: The absence of crystals (ruling out gout or pseudogout) and no organisms or WBCs (ruling out septic arthritis) suggests a non-infectious etiology.
    \item Risk Factors: This is commonly associated with diabetes mellitus or other neuropathic conditions, which should be explored in this patient.
\end{enumerate}

\textit{Research completed in 10m · 20 sources · 75 searches}

\textbf{Why Charcot Arthropathy Fits Best}

Several clinical clues in this case strongly support Charcot neuropathic arthropathy as the correct diagnosis:
\begin{itemize}
    \item \textbf{Normal X-rays despite severe symptoms:} In early Charcot arthropathy (Stage 0), patients have an acutely inflamed joint before any bony destruction is visible on radiographs. This patient’s X-ray showed no acute abnormalities even though his ankle was red, hot, and swollen – a classic scenario for early Charcot changes...
    \item \textbf{Peripheral neuropathy signs and risk factors:} Charcot arthropathy virtually always occurs in the setting of neuropathy. Although not explicitly stated, a 68-year-old patient is likely to have neuropathy...
    \item \textbf{Laboratory and fluid studies exclude other causes:} The synovial fluid had \textbf{no crystals, no organisms, and minimal inflammatory cells,} which is exactly what one might find in Charcot...
    \item \textbf{Episode triggered by activity (trauma):} The history that symptoms recur “following activity” aligns with Charcot pathogenesis. Episodes often start after minor trauma or increased weight-bearing exercise...
    \item \textbf{Initial improvement with NSAIDs, then worsening on steroids:} In Charcot arthropathy, treating with just anti-inflammatories can temporarily reduce swelling and pain, but if the patient continues to walk on the joint, the underlying bony collapse progresses...
\end{itemize}

In summary, the constellation of a relatively \textbf{painless, swollen red ankle}, with \textbf{normal X-rays}, \textbf{negative joint fluid}, and a history of neuropathy or diabetes, is virtually pathognomonic for \textbf{Charcot neuropathic arthropathy} once infection and crystals are ruled out. The case matches this picture closely, making Charcot joint the most likely diagnosis.

\textbf{Sources:}
\begin{enumerate}[label=\arabic*., itemsep=0pt, topsep=4pt]
    \item Amelia Harris, Melanie Violand. Charcot Neuropathic Osteoarthropathy. StatPearls, 2023
    \item Botek G, Hild G. Differential diagnosis of Charcot arthropathy. Lower Extremity Review. 2011
    \item Pinzur MS. Charcot foot: clinical clues, diagnostic strategies, and treatment principles. American Family Physician. 2018;97(9):594-600
    \item Rosskopf AB, et al. The Charcot foot: a pictorial review. Insights Imaging. 2019;10(1):77
    \item Singh D, et al. Diagnosing acute monoarthritis in adults: a practical approach. Am Fam Physician. 2003;68(1):83-90
    \item Lower Extremity Review (2011). Charcot arthropathy imaging and management
    \item Radiopaedia. Osteoarthritis of the ankle – radiographic features
    \item Lower Extremity Review (2011). Charcot foot vs. gout and infection
\end{enumerate}
\end{competitorbox}

\vspace{1.5em} 

\begin{ourmodelbox}{\textcolor{white}{\faCheckCircle\ \ Our Model: AgentFrontier-30B-A3B (Correct)}}

\begin{roundbox}{Round 1: Initial Analysis \& Planning}
\paragraph{Status Report and Deep Analysis}
\textbf{Problem Essence Analysis:} The patient is a 68-year-old presenting with acute ankle pain, swelling, erythema, and tenderness following a long walk... The paradoxical worsening with prednisone is particularly notable and may point to a condition where steroids exacerbate symptoms, such as certain infections or rare inflammatory disorders...

\textbf{Problem-Solving Planning:} The analysis confirms that common diagnoses are unlikely. However, chronic osteomyelitis or rare conditions like tuberculous arthritis remain possibilities... Further verification through academic sources is needed to identify conditions that match this specific presentation, especially those where prednisone worsens symptoms.

\vspace{0.5em}
\noindent\textbf{Tool Call:}
\begin{lstlisting}[style=jsonstyle]
{"name": "google_scholar", "arguments": {"query": ["ankle pain swelling erythema no crystals no organisms elevated uric acid", "differential diagnosis ankle arthritis negative synovial fluid", "prednisone worsens arthritis no crystals"]}}
\end{lstlisting}
\end{roundbox}
\vspace{1em}

\begin{roundbox}{Round 2: Synthesizing Evidence \& Refining Hypothesis}
\paragraph{Status Report and Deep Analysis}
\textbf{Progress Achieved:} Based on the initial clinical presentation and subsequent diagnostic findings, the following comprehensive information has been collected...
\begin{itemize}[leftmargin=*, topsep=3pt, itemsep=0pt]
    \item \textbf{Differential Diagnosis Analysis}: Osteoarthritis (Unlikely), Charcot Arthropathy (Unlikely due to no mention of neuropathy), Septic Arthritis (Ruled out), Pseudogout (Ruled out), \textbf{Chronic Osteomyelitis (Possible)}.
    \item \textbf{Key Diagnostic Challenges}: Paradoxical worsening with prednisone therapy suggests an infection that is exacerbated by immunosuppression...
    \item \textbf{Literature Findings from Tool Response}: Search results indicate that certain infections (particularly tuberculous or fungal arthritis) can present with negative synovial fluid cultures and gram stains... Chronic osteomyelitis can present with intermittent symptoms and minimal radiographic changes.
\end{itemize}

\textbf{Next Steps Plan:} Given the complex presentation... the next step should focus on obtaining more specific diagnostic information about chronic osteomyelitis presentations and its differentiation from other conditions.

\vspace{0.5em}
\noindent\textbf{Tool Call:}
\begin{lstlisting}[style=jsonstyle]
{"name": "google_scholar", "arguments": {"query": ["chronic osteomyelitis ankle differential diagnosis", "osteomyelitis ankle negative synovial fluid", "ankle arthritis steroid exacerbation", "tuberculous arthritis ankle presentation"]}}
\end{lstlisting}
\end{roundbox}
\vspace{1em}

\begin{roundbox}{Round 3: Finalizing Diagnosis}
\paragraph{Status Report and Deep Analysis}
\textbf{Progress Achieved:}
The second, more targeted literature search provided crucial confirmatory evidence, allowing for a final synthesis of all findings and a resolution of the remaining diagnostic uncertainties.
\begin{itemize}[leftmargin=*, topsep=3pt, itemsep=0pt]
\item \textbf{Synthesized Literature Findings}: The model integrated results from the second tool call to build a robust case for chronic osteomyelitis.
\begin{itemize}[leftmargin=*, topsep=2pt, itemsep=0pt]
\item \textbf{Explaining Negative Findings}: The search for "osteomyelitis ankle negative synovial fluid" yielded articles confirming that false negative results in synovial fluid cultures and gram stains are possible, and that "culture-negative osteomyelitis" is a recognized clinical entity. This directly addressed a key uncertainty from Round 1, explaining why a bone infection could be present despite "clean" joint fluid. Similarly, literature confirmed that initial radiographs are often negative in osteomyelitis, validating the patient's repeated negative X-rays.
\item \textbf{Confirming the Steroid Paradox}: The search on "ankle arthritis steroid exacerbation" reinforced the understanding that corticosteroid use can lead to adverse outcomes. By cross-referencing this with the infectious nature of osteomyelitis, the model concluded that the patient's worsening symptoms were a classic sign of an occult infection being unmasked by immunosuppression, transforming this paradoxical finding into a powerful diagnostic clue.
\item \textbf{Differentiating from Alternatives}: The search on "tuberculous arthritis ankle presentation" revealed that it typically presents as a subacute or chronic monoarthritis, often over weeks to months. This contrasts with the patient's more acute onset following a specific event (the long walk), making tuberculous arthritis a less likely, though still possible, alternative compared to chronic osteomyelitis.
\end{itemize}
\item \textbf{Re-evaluation of Clinical Data}: In light of the new evidence, the model re-evaluated the initial data with greater clarity.
\begin{itemize}[leftmargin=*, topsep=2pt, itemsep=0pt]
\item The \textbf{bony tenderness} is now seen as a specific sign of bone involvement, not just generalized joint pain.
\item The elevated \textbf{C-reactive protein (CRP)} is confirmed as a marker of significant inflammation, consistent with a bone infection.
\item The slightly elevated \textbf{uric acid} is confidently dismissed as a non-specific finding or a red herring, as no other evidence supports a diagnosis of gout.
\end{itemize}
\item \textbf{Diagnostic Certainty Assessment}: The model's confidence in the diagnosis of Chronic Osteomyelitis increased substantially. All other possible diseases had been systematically ruled out by clinical data. Chronic Osteomyelitis remained the only diagnosis that could cohesively explain the entire constellation of findings: persistent symptoms, bony tenderness, negative initial imaging and fluid analysis, and the critical paradoxical response to steroid therapy. The supporting literature provided a solid foundation for this conclusion.
\end{itemize}

\textbf{Next Steps Plan:} No further investigation is required. The diagnostic pathway has been conclusively established through systematic elimination of alternatives and the integration of clinical findings with targeted, confirmatory literature evidence. The available information is sufficient to provide a definitive diagnosis.
\end{roundbox}

\vspace{1em} 

\begin{answerbox}
Based on the comprehensive clinical evaluation and integration of supporting literature evidence, this patient's diagnosis is \textcolor{mygreen}{\textbf{Chronic Osteomyelitis}}.

\subsection*{Diagnostic Process}
The diagnostic journey for this 68-year-old patient began with acute ankle pain and swelling... The most critical diagnostic turning point came with the synovial fluid analysis... combined with the patient's paradoxical worsening during prednisone therapy...

\subsection*{Differential Diagnosis Elimination}
Through systematic exclusion of alternative diagnoses, chronic osteomyelitis emerges as the most consistent explanation...

\subsection*{Pathophysiological Mechanisms Supporting Chronic Osteomyelitis}
Chronic osteomyelitis provides the most coherent explanation for this clinical scenario through several key mechanisms:
\paragraph{Insidious Presentation with Normal Radiographs} Chronic osteomyelitis often presents with minimal early radiographic changes...
\paragraph{Paradoxical Response to Steroids} The worsening during prednisone therapy is particularly telling. Corticosteroids suppress the immune system, allowing previously contained bacterial colonies within necrotic bone to proliferate unchecked...
\paragraph{Synovial Fluid Findings} The negative synovial fluid analysis does not exclude osteomyelitis, as the infection originates in the bone marrow...
\paragraph{Bony Tenderness Without Acute Trauma} The mild bony tenderness... indicates underlying bone pathology consistent with osteomyelitis...

\subsection*{Supporting Evidence from Literature}
Recent medical literature corroborates this diagnostic conclusion. Studies indicate that chronic osteomyelitis of the foot and ankle frequently presents with challenging diagnostic features... The paradoxical worsening with corticosteroid therapy aligns perfectly with the known pathophysiology...

\subsection*{Conclusion}
In summary, this patient's clinical presentation—characterized by acute inflammatory symptoms, normal radiographs, negative synovial fluid analysis, and paradoxical worsening with steroid therapy—best aligns with chronic osteomyelitis rather than any of the other listed conditions...
\end{answerbox}

\end{ourmodelbox}

\newpage
\section{Prompts Used in Experiments}
\label{app:prompt}
The key prompts used in our experiments are presented below to ensure reproducibility.
\subsection{Evaluation Prompt}
\label{subsec:eval_prompt}

\begin{tcolorbox}[
    colback=black!5!white,  
    colframe=black!75!black, 
    title=\textbf{Evaluation Prompt},
    fonttitle=\bfseries,
    boxrule=0.5pt,
    arc=2mm,
]
Judge whether the following \texttt{[response]} to \texttt{[question]} is correct or not based on the precise and unambiguous \texttt{[correct\_answer]} below.

\vspace{1em} 

\textbf{[question]}: \texttt{\{question\}}

\textbf{[response]}: \texttt{\{response\}}

\vspace{1em}

Your judgement must be in the format and criteria specified below:

\begin{description}
    \item[\texttt{extracted\_final\_answer:}] The final exact answer extracted from the \texttt{[response]}. Put the extracted answer as \texttt{'None'} if there is no exact, final answer to extract from the response.

    \item[\texttt{[correct\_answer]:}] \texttt{\{correct\_answer\}}

    \item[\texttt{reasoning:}] Explain why the \texttt{extracted\_final\_answer} is correct or incorrect based on \texttt{[correct\_answer]}, focusing only on if there are meaningful differences between \texttt{[correct\_answer]} and the \texttt{extracted\_final\_answer}. Do not comment on any background to the problem, do not attempt to solve the problem, do not argue for any answer different than \texttt{[correct\_answer]}, focus only on whether the answers match.

    \item[\texttt{correct:}] Answer \texttt{'yes'} if \texttt{extracted\_final\_answer} matches the \texttt{[correct\_answer]} given above, or is within a small margin of error for numerical problems. Answer \texttt{'no'} otherwise, i.e. if there if there is any inconsistency, ambiguity, non-equivalency, or if the extracted answer is incorrect.

    \item[\texttt{confidence:}] The extracted confidence score between \texttt{0|\%|} and \texttt{100|\%|} from \texttt{[response]}. Put \texttt{100} if there is no confidence score available.
\end{description}
\end{tcolorbox}

\subsection{Similarity Filter Prompt}
\label{subsec:sim_prompt}

\begin{tcolorbox}[
    colback=black!5!white,
    colframe=black!75!black,
    title=\textbf{Similarity Filter Prompt},
    fonttitle=\bfseries,
    boxrule=0.5pt,
    arc=2mm,
]
Determine if the candidate QA pair expresses \textbf{EXACTLY} the same specific question and answer as the reference QA pair.

\paragraph{Requirements:}
\begin{enumerate}
    \item The question must ask for identical information with identical technical requirements.
    \item The answer must provide identical content with identical technical details.
    \item Any difference in the specific information requested or provided means they are NOT identical.
    \item Pay special attention to mathematical expressions, symbols, and technical specifications.
\end{enumerate}
\end{tcolorbox}

\subsection{Agentic Refinement Prompt}

\begin{tcolorbox}[
  colback=black!5!white,
  colframe=black!75!black,
  fonttitle=\bfseries,
  title=Prompt for Agentic Refinement ($\mathcal{A}_{\text{refine}}$),
  width=\textwidth
]

\textbf{Role and Objective:} \\
You are a sophisticated agent tasked with iterative data refinement. Your primary mission is to transform a given Question-Answer pair $(q_k, a_k)$ into a more complex, in-depth, and factually grounded pair $(q_{k+1}, a_{k+1})$. This escalation must be achieved by leveraging a specialized tool suite $\mathcal{T} = \{T_{\text{search}}, T_{\text{scholar}}, T_{\text{browser}}, T_{\text{code}}\}$.

\vspace{2mm}
\textbf{Input:} \\
The current QA pair QA pair $(q_k, a_k)$ in a structured format.

\vspace{2mm}
\textbf{Mandatory Refinement Protocol:} \\
Your task is to generate a new, superior QA pair by applying one or more of the following four refinement dimensions. For each generated pair, you \textbf{must} utilize the provided tools and explicitly log their usage.

\begin{enumerate}
    \item \textbf{Knowledge Expansion:}
    \begin{itemize}
        \item \textbf{Objective:} Broaden the informational scope of the QA pair.
        \item \textbf{Action:} You \textbf{must} use the $T_{\text{search}}$, $T_{\text{scholar}}$, or $T_{\text{browser}}$ tools to discover and retrieve relevant background knowledge, historical context, or contrasting perspectives.
        \item \textbf{Implementation:} Weave this new information seamlessly into the refined question ($q_{k+1}$) and provide a comprehensive explanation in the refined answer ($a_{k+1}$).
    \end{itemize}

    \item \textbf{Conceptual Abstraction:}
    \begin{itemize}
        \item \textbf{Objective:} Elevate the level of abstract reasoning required.
        \item \textbf{Action:} Analyze the core concepts within $(q_k, a_k)$. Formulate a new question ($q_{k+1}$) that requires identifying higher-level principles, synthesizing information to uncover subtle relationships, or drawing non-obvious analogies.
        \item \textbf{Implementation:} The refined answer ($a_{k+1}$) must explicitly articulate this abstract principle or relationship. You may use $T_{\text{scholar}}$ to find established theoretical frameworks to aid this process.
    \end{itemize}

    \item \textbf{Factual Grounding:}
    \begin{itemize}
        \item \textbf{Objective:} Enhance the factual accuracy, precision, and verifiability.
        \item \textbf{Action:} You \textbf{must} use $T_{\text{search}}$ and $T_{\text{scholar}}$ to perform multi-source cross-validation of the facts and claims in $a_k$.
        \item \textbf{Implementation:} Augment the refined answer ($a_{k+1}$) with precise quantitative data, specific named entities, and direct citations or references to the authoritative sources you retrieved.
    \end{itemize}

    \item \textbf{Computational Formulation:}
    \begin{itemize}
        \item \textbf{Objective:} Introduce a verifiable computational or logical reasoning challenge.
        \item \textbf{Action:} You \textbf{must} use the $T_{\text{code}}$ tool (a Python execution environment) to design a new question ($q_{k+1}$) that necessitates a quantitative calculation or algorithmic simulation.
        \item \textbf{Implementation:} The refined answer ($a_{k+1}$) must contain: (1) The complete, executable Python code block used to solve the problem, and (2) The final output produced by the code, along with a brief explanation.
    \end{itemize}
\end{enumerate}

\textbf{Tool Usage Protocol:} \{tools\}\\
\textbf{Final Instruction:} \\
Proceed with the refinement of the provided $(q_k, a_k)$. Your response must be only the final JSON object.
\end{tcolorbox}

\clearpage
\bibliography{biblio}
\bibliographystyle{colm2024_conference}

\end{document}